\pdfoutput=1

\documentclass[11pt,table,xcdraw]{article}

\usepackage[preprint]{coling}

\usepackage{times}
\usepackage{latexsym}

\usepackage[T1]{fontenc}
\usepackage[utf8]{inputenc}
\usepackage{microtype}
\usepackage{adjustbox}
\usepackage{inconsolata}
\usepackage{booktabs}
\usepackage{graphicx}
\usepackage[inline]{enumitem}
\usepackage{tipa}
\usepackage{fontawesome5}

\usepackage{amsmath,amssymb}

\usepackage{pgfplots, pgfplotstable}
\pgfplotsset{compat=1.15}
\usetikzlibrary{positioning}

\newcommand{\horizontalspacer}{\\[-8pt]}

\newcommand{\ortho}[1]{$\langle$\texttt{#1}$\rangle$}

\title{Add Noise, Tasks, or Layers? MaiNLP at the VarDial 2025 Shared Task on Norwegian Dialectal Slot and Intent Detection}

\author{Verena Blaschke*
\And Felicia Körner* \\
MaiNLP, Center for Information and Language Processing, LMU Munich, Germany \\
Munich Center for Machine Learning (MCML), Munich, Germany \\ {\tt \{verena.blaschke, f.koerner, b.plank\}@lmu.de} 
\And Barbara Plank}

\newcommand\blfootnote[1]{%
  \begingroup
  \renewcommand\thefootnote{}\footnote{#1}%
  \addtocounter{footnote}{-1}%
  \endgroup
}

\newcommand{\skiplarge}{\hspace{10pt}}
\newcommand{\skipmid}{\hspace{7pt}}
\newcommand{\skipsmall}{\,}
\newcommand{\skiptiny}{\kern1pt}

\newcommand{\multitask}{$\times$}
\newcommand{\sequential}{$\rightarrow$}
\newcommand{\fonescore}{F\textsubscript{1}}

\begin{document}
\maketitle

\begin{abstract}
Slot and intent detection (SID) is a classic natural language understanding task. 
Despite this, research has only more recently begun focusing on SID for dialectal and colloquial varieties.
Many approaches for low-resource scenarios have not yet been applied to dialectal SID data, or compared to each other on the same datasets.
We participate in the VarDial 2025 shared task on slot and intent detection in Norwegian varieties, and compare multiple set-ups: varying the training data (English, Norwegian, or dialectal Norwegian), injecting character-level noise, training on auxiliary tasks, and applying Layer Swapping, a technique in which layers of models fine-tuned on different datasets are assembled into a model.
We find noise injection to be beneficial while the effects of auxiliary tasks are mixed.
Though some experimentation was required to successfully assemble a model from layers, it worked surprisingly well; a combination of models trained on English and small amounts of dialectal data produced the most robust slot predictions.
Our best models achieve 97.6\% intent accuracy and 85.6\% slot \fonescore{} in the shared task.
\end{abstract}
\blfootnote{*Equal contribution.}%

\section{Introduction}
\label{sec:intro}
Slot and intent detection (SID) is a classic natural language understanding (NLU) task. 
Research today has mainly focused on standard languages with many speakers~\cite[e.g.,][]{schuster-etal-2019-cross-lingual, xu-etal-2020-end, li-etal-2021-mtop,fitzgerald-etal-2023-massive}.
However, even when performance on a related standard language is high, SID for non-standard varieties can be challenging. This can be due to spelling variation \cite{srivastava-chiang-2023-fine} and syntactic differences that complicate cross-lingual slot filling \cite{artemova-etal-2024-exploring}. Furthermore, the lack of task data in the relevant language varieties complicates the adaptation of SID models to under-resourced varieties.

\newcommand{\colournorsid}{red!30!yellow!50}
\newcommand{\colourbokmaal}{red!20}
\newcommand{\colournorwegian}{red!60!yellow!40}
\newcommand{\coloureng}{green!20!blue!30}
\newcommand{\notestyle}[1]{\footnotesize\textcolor{black!40}{#1}}
\newlength{\nodeheight}
\setlength{\nodeheight}{16pt}
\newlength{\anchordist}
\setlength{\anchordist}{24pt}
\newlength{\mediumanchordist}
\setlength{\mediumanchordist}{29pt}
\newlength{\largeanchordist}
\setlength{\largeanchordist}{34pt}
\newlength{\xlargeanchordist}
\setlength{\xlargeanchordist}{44pt}
\newlength{\largeandonepointfiveanchordist}
\setlength{\largeandonepointfiveanchordist}{46pt}
\newlength{\onepointfiveanchordist}
\setlength{\onepointfiveanchordist}{36pt}
\newlength{\halfanchordist}
\setlength{\halfanchordist}{12pt}
\newlength{\noteanchordist}
\setlength{\noteanchordist}{18pt}
\newlength{\seclabeldist}
\setlength{\seclabeldist}{12pt}

\begin{figure}[t]
\centering
\adjustbox{max width=\linewidth}{%
\begin{tikzpicture}[
    align=left,
    mynode/.style={draw=none, inner sep=2pt, outer sep=0pt, minimum size=\nodeheight, rounded corners=5pt, font=\ttfamily, align=center},
    nordial/.style={mynode, fill=\colournorsid},
    bokmaal/.style={mynode, fill=\colourbokmaal},
    english/.style={mynode, fill=\coloureng},
    plm/.style={mynode, fill=black!20},
    mydashed/.style={dashed, black!30, line width=1pt},
    seclabelnode/.style={draw=none, align=center, rotate=90},
]
\node[plm] at (0, 2) (PLM1) {\,PLM\,};
\node[plm, below=\mediumanchordist of PLM1.west, anchor=west] (PLM2) {\,PLM\,};
\node[plm, below=\mediumanchordist of PLM2.west, anchor=west] (PLM3) {\,PLM\,};
\node[english] at (2, 2) (ENGSID1) {\,English SID\,};
\node[below=\noteanchordist of ENGSID1.west, anchor=west]
    {\notestyle{\texttt{Remind me to pick up bread today}}\\[-6pt]\,};
\node[bokmaal, below=\mediumanchordist of ENGSID1.west, anchor=west] (BOKSID1) {\,MT'ed Norw.\ SID\,};
\node[below=\noteanchordist of BOKSID1.west, anchor=west]
    {\notestyle{\texttt{Minner} \textit{(sic)} \texttt{meg på å hente brød i dag}}\\[-6pt]\,};
\node[nordial, below=\mediumanchordist of BOKSID1.west, anchor=west] (NORSID1) {\,NorSID (dev)\,};
\node[below=\noteanchordist of NORSID1.west, anchor=west]
    {\notestyle{\texttt{Minn mæ på å send inn mine timeplana}}\\[-6pt]
    \notestyle{`Remind me to submit my lesson plans'}\,};
\node[nordial] at (7.5, 2) (EVAL1) {\,NorSID\,};
\node[nordial, below=\mediumanchordist of EVAL1.west, anchor=west] (EVAL2) {\,NorSID\,};
\node[nordial, below=\mediumanchordist of EVAL2.west, anchor=west] (EVAL3) {\,NorSID\,};
\draw[->] (PLM1) -- (ENGSID1);
\draw[->] (ENGSID1) -- (EVAL1);
\draw[->] (PLM2) -- (BOKSID1);
\draw[->] (BOKSID1) -- (EVAL2);
\draw[->] (PLM3) -- (NORSID1);
\draw[->] (NORSID1) -- (EVAL3);

\node[plm, below=\xlargeanchordist of PLM3.west, anchor=west] (PLM4) {\,PLM\,};
\node[bokmaal, below=\xlargeanchordist of NORSID1.west, anchor=west] (NOISED) {\,MT'ed Norw.\ SID with noise\,};
\node[below=\noteanchordist of NOISED.west, anchor=west]
    {\notestyle{\texttt{Mi\textcolor{black!40!red}{\textbf{d}}ner meg på å hen\textcolor{black!40!red}{\textbf{,}}e brød i dag}}\\[-6pt]\,};
\node[nordial, below=\xlargeanchordist of EVAL3.west, anchor=west] (EVAL4) {\,NorSID\,};
\draw[->] (PLM4) -- (NOISED);
\draw[->] (NOISED) -- (EVAL4);

\node[plm, below=\largeandonepointfiveanchordist of PLM4.west, anchor=west] (PLM5) {\,PLM\,};
\node[plm, below=\onepointfiveanchordist of PLM5.west, anchor=west] (PLM6) {\,PLM\,};
\node[plm, below=\anchordist of PLM6.west, anchor=west] (PLM7) {\,PLM\,};
\node[plm, top color=\colournorsid, bottom color=\colourbokmaal, below=\largeandonepointfiveanchordist of NOISED.west, anchor=west] (AUX1) {\,Aux.\ task\,};
\node[align=center, below=\noteanchordist of AUX1.center, anchor=center]
    {\notestyle{\textit{Dialect ID, POS,}}\\[-6pt]
    \notestyle{\textit{parsing or NER}}\,};
\node[english, below=\largeanchordist of NOISED.center, anchor=west] (ENGSID2) {\,English SID\,};
\node[bokmaal, below=\anchordist of ENGSID2.west, anchor=west] (BOKSID2) {\,MT'ed Norw.\ SID\,};
\node[plm, top color=\colournorwegian, bottom color=\coloureng, below=\onepointfiveanchordist of AUX1.west, anchor=west] (AUX2) {\,Aux.\ task + English SID\,};
\node[plm, top color=\colournorwegian, bottom color=\colourbokmaal, below=\anchordist of AUX2.west, anchor=west] (AUX3) {\,Aux.\ task + MT'ed Norw.\ SID\,};
\node[nordial, below=\largeanchordist of EVAL4.west, anchor=west] (EVAL5) {\,NorSID\,};
\node[nordial, below=\anchordist of EVAL5.west, anchor=west] (EVAL6) {\,NorSID\,};
\node[nordial, below=\anchordist of EVAL6.west, anchor=west] (EVAL7) {\,NorSID\,};
\node[nordial, below=\anchordist of EVAL7.west, anchor=west] (EVAL8) {\,NorSID\,};
\draw[->] (PLM5) -- (AUX1);
\draw[->] (AUX1.east) -- (ENGSID2.west);
\draw[->] (ENGSID2) -- (EVAL5);
\draw[->] (AUX1.east) -- (BOKSID2.west);
\draw[->] (BOKSID2) -- (EVAL6);
\draw[->] (PLM6) -- (AUX2);
\draw[->] (AUX2) -- (EVAL7);
\draw[->] (PLM7) -- (AUX3);
\draw[->] (AUX3) -- (EVAL8);

\node[plm, below=\largeanchordist of PLM7.west, anchor=west] (PLM8) {\,PLM\,};
\node[plm, below=\anchordist of PLM8.west, anchor=west] (PLM9) {\,PLM\,};
\node[english, below=\largeanchordist of AUX3.west, anchor=west] (ENGSID3) {\,English SID\,};
\node[nordial, below=\anchordist of ENGSID3.west, anchor=west] (NORSID2) {\,NorSID (dev)\,};
\node[nordial, below=\largeanchordist of EVAL8.west, anchor=west] (EVAL9) {\,NorSID\,};
\node[font=\itshape, above=\halfanchordist of NORSID2.east, anchor=west] (INSERT) {\,replace layers\,};
\coordinate[above=\halfanchordist of INSERT.east, anchor=west] (ARROWJOIN);
\coordinate[below=\halfanchordist of PLM9.west, anchor=west] (PLMLAST);
\draw[->] (PLM8) -- (ENGSID3);
\draw[->] (ENGSID3) -- (EVAL9);
\draw[->] (PLM9) -- (NORSID2);
\draw[->] (NORSID2.east) -- (INSERT) -- (ARROWJOIN);

\node[] at (4, 2.8) {\,Finetuning\,};
\node[] at (7.5, 2.8) {\,Eval\,};
\node[seclabelnode, left=\seclabeldist of PLM2.west, anchor=center] {\S\ref{sec:methods-baselines}\\[-4pt]Baselines};
\node[seclabelnode, left=\seclabeldist of PLM4.west, anchor=center] {\S\ref{sec:methods-noise}\\[-4pt]Noise};
\node[seclabelnode, left=\seclabeldist of PLM7.west, anchor=west] {\S\ref{sec:methods-auxtasks}\\[-4pt]Auxiliary tasks};
\node[seclabelnode, left=\seclabeldist of PLMLAST.west, anchor=west, xshift=-15pt] {\S\ref{sec:methods-layer-swapping}\\[-4pt]Layer Swapping};
\end{tikzpicture}
}
\caption{\textbf{Overview of our approaches:} pre-trained language models (PLMs) fine-tuned on English, machine-translated Norwegian data or the dialectal development set; noise injection into the Norwegian data; training on auxiliary tasks in addition to SID data (sequentially or jointly); assembling layers of models fine-tuned on different datasets.}
\label{fig:overview-approaches}
\end{figure}
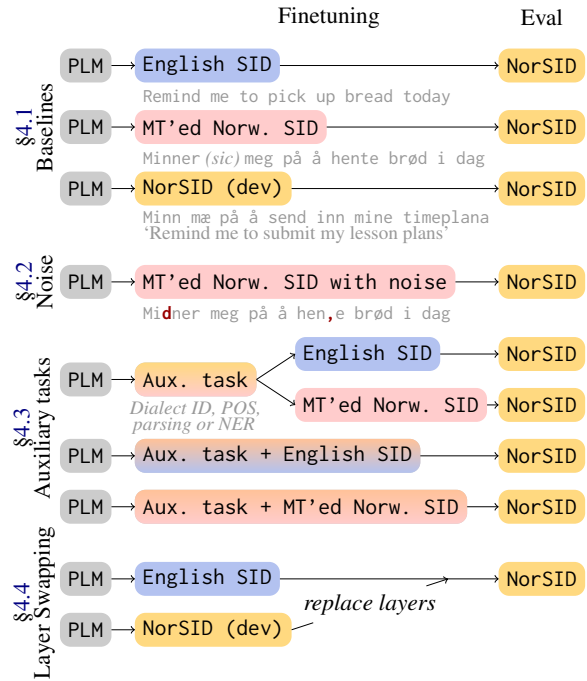

In this paper, we report on the results of our participation in the VarDial 2025 shared task on slot and intent detection in Norwegian standard and dialect varieties (NorSID; \citealp{scherrer-etal-2025-vardial}). We compare multiple strategies for improving the performance of SID systems  (Figure~\ref{fig:overview-approaches}):
\begin{enumerate}
    \item Fine-tuning models on large amounts of gold-standard English or silver-standard Norwegian data, or smaller amounts of gold-standard Norwegian dialect data~(\S\ref{sec:methods-baselines});
    \item Adding noise to the Norwegian training data to make models more robust to spelling variation~(\S\ref{sec:methods-noise});
    \item Additionally training on auxiliary NLP tasks in Norwegian~(\S\ref{sec:methods-auxtasks});
    \item Assembling layers of models fine-tuned on different tasks or languages into a new model to combine their capabilities~(\S\ref{sec:methods-layer-swapping}).
\end{enumerate}

\noindent
We share our code at \url{https://github.com/mainlp/NorSID}.

\section{Related Work}
\label{sec:related-work}

In the past few years, research on SID for dialects and non-standard languages has gained popularity.
The multilingual SID dataset xSID \cite{van-der-goot-etal-2021-masked, aepli-etal-2023-findings, winkler-etal-2024-slot} contains evaluation data in over a dozen languages, including non-standard varieties like Neapolitan, and German dialects spoken in Switzerland, South Tyrol, and Bavaria.
It has recently been extended with translations into Norwegian dialects \citep{maehlum-scherrer-2024-nomusic}, which are the focus of this shared task.
We provide more details in~\S\ref{sec:data}.

Using xSID, \citet{van-der-goot-etal-2021-masked} investigate multi-task learning with auxiliary tasks in the target language (or a closely related standard language). 
Similarly, \citet{krueckl2025improving} include auxiliary tasks in multi-task learning and intermediate-task training set-ups for dialectal SID.
Both studies find that the effects depend on both the auxiliary task(s) and the target task.
We include auxiliary tasks in one of our experiments~(\S\ref{sec:methods-auxtasks}).

Two previous shared tasks have focused on dialectal SID \cite{aepli-etal-2023-findings, malaysha-etal-2024-arafinnlp}.
Useful approaches by the participants were to train on SID data in multiple languages \cite{kwon-etal-2023-sidlr}, injecting character-level noise into the training data (\citealp{srivastava-chiang-2023-fine}; we use a similar method in~\S\ref{sec:methods-noise}), and ensembling models trained on dialectal translations of the training set \cite{ramadan-etal-2024-ma, elkordi-etal-2024-alexunlp24, fares-touileb-2024-babelbot}.

Outside the context of a shared task, \citet{abboud-oz-2024-towards} also focus on generating synthetic dialectal training data.
Lastly, \citet{munoz-ortiz-etal-2025-evaluating} find that visual input representations are more robust than subword token embeddings when transferring German intent classification models to related dialects.

Numerous other methods for improving NLP performance in low-resource settings exist \cite{hedderich-etal-2021-survey}, many of which have not yet been applied to dialectal or cross-lingual SID.
One recently proposed approach is assembling layers of models trained on different tasks or languages into a new model \citep{bandarkar2024layerswappingzeroshotcrosslingual}, which we explore in~\S\ref{sec:methods-layer-swapping}.
\section{Data}
\label{sec:data}
\begin{table}
\adjustbox{max width=\linewidth}{%
\begin{tabular}{@{}l@{~~}rr@{~~}r@{}}
\toprule
\textbf{Label type / data subset} & \textbf{Train} & \textbf{Dev} & \textbf{Test} \\ 
\midrule
Intents & \multicolumn{1}{r}{18} & 15 & 15 \\
Slot types & \multicolumn{1}{r}{40} & 33 & 34 \\ 
\midrule
\midrule
English & 43k & \multicolumn{2}{c}{(not used)} \\
Bokmål & \llap{(MT)} 43k & 1$\times$300 & 1$\times$500 \\
North Norwegian & --- & 2$\times$300 & 2$\times$500 \\
Trønder Norwegian & --- & 3$\times$300 & 3$\times$500 \\
West Norwegian & --- & 5$\times$300 & 5$\times$500 \\
\midrule
Total (evaluation) & --- & 3\,300 & 5\,500 \\
Training on dev & 2\,970 & 330 & 5\,500 \\
\bottomrule
\end{tabular}
}
\caption{\textbf{Distribution of labels and languages/\allowbreak{}dialects in the data.}
While 15 intent types occur in both the development and test splits, only 14 of them overlap.
}
\label{tab:xsid-stats}
\end{table}

We use the xSID~0.6 dataset \cite{van-der-goot-etal-2021-masked} and its Norwegian extension NoMusic \cite{maehlum-scherrer-2024-nomusic}.
xSID combines re-annotated versions of two SID datasets \cite{coucke2018snips, schuster-etal-2019-cross-lingual}.
It includes 43k~English training sentences, as well as smaller development and test datasets that have been translated into other languages.
The shared task also includes an automatic translation of the training set into Bokmål \cite{scherrer-etal-2025-vardial}.
For these sentence-level translations, the intent labels remained unchanged, while the slot annotations were automatically projected during the translation.

NoMusic provides translations of xSID's development and test utterances into the Norwegian Bokmål orthography and Norwegian dialects from three of the four major dialect groups (Table~\ref{tab:xsid-stats}).
The dialect groups have two to five different translations each.
In some of our experiments, we train on the development set, which we split into a training and new development set according to a 90:10 ratio. 

Slots are annotated in the \texttt{BIO} scheme.
Intent classification is measured with accuracy, and slot filling with strict span \fonescore{}.

One of our approaches uses datasets for auxiliary tasks; these are described in~\S\ref{sec:methods-auxtasks}. 
We also use additional datasets for Layer Swapping experiments, described in~\S\ref{sec:methods-layer-swapping}.

\section{Methodology}
\label{sec:methods}
We construct several baselines that differ in their training data and pretrained language model (PLM) choices~(\S\ref{sec:methods-baselines}).
We subsequently build on (some of) these baselines to examine the effects of different recent approaches to improving performance on low-resource language data. 
We submitted three systems each for intent classification and for slot filling; here we also discuss %
models we tested but did not submit to the shared task.
When selecting systems for submission, we considered their performance on the development set while also aiming for a diverse set of systems.

We use MaChAmp (\citealp{van-der-goot-etal-2021-massive}; v.0.4.2, commit \texttt{052044a}) with default hyperparameters to fine-tune the PLMs to simultaneously predict slot and intent labels.
The slot predictions are decoded via a conditional random field (CRF;  \citealp{lafferty2001conditional}).
Each model is fine-tuned for 20~epochs, and the best epoch is chosen based on performance on the development set. 
For Layer Swapping, we build on an implementation of JointBERT~(\S\ref{sec:methods-layer-swapping}).

\subsection{Baselines}
\label{sec:methods-baselines}

We fine-tune three PLMs as baselines:
\begin{enumerate*}[label=\itshape\arabic*)]
\item
the monolingual Norwegian NorBERT~v3 \cite{samuel-etal-2023-norbench};\footnote{\href{https://huggingface.co/ltg/norbert3-base}{\texttt{ltg/norbert3-base}} (\href{https://huggingface.co/datasets/choosealicense/licenses/blob/main/markdown/apache-2.0.md}{Apache~2.0})}
\item
ScandiBERT \cite{snaebjarnarson-etal-2023-transfer},%
\footnote{\href{https://huggingface.co/vesteinn/ScandiBERT}{\texttt{vesteinn/ScandiBERT}} (\href{https://huggingface.co/datasets/choosealicense/licenses/blob/main/markdown/agpl-3.0.md}{AGPL~3.0})}
which was pretrained on data in Norwegian, Danish, Swedish, Icelandic, and Faroese;
and
\item
mDeBERTa~v3 \cite{he2021deberta,he2023debertav3},\footnote{\href{https://huggingface.co/microsoft/mdeberta-v3-base}{\texttt{microsoft/mdeberta-v3-base}} (\href{https://huggingface.co/datasets/choosealicense/licenses/blob/main/markdown/mit.md}{MIT})}
which was pretrained on 100 languages, including Norwegian \cite{conneau-etal-2020-unsupervised}, and has performed well on dialectal SID data \cite{artemova-etal-2024-exploring, krueckl2025improving}.
\end{enumerate*}

We fine-tune each PLM three times: once on xSID's English training data, once on the machine-translated Norwegian version, and once on NoMusic's development data.

\paragraph{Shared task submission}
We include the mDeBERTa model trained on the development data in our submissions (slots and intents).\footnote{\texttt{mainlp\_\{slots,intents\}1\_mdeberta\_siddial\_\allowbreak{}8446}}

\subsection{Character-Level Noise}
\label{sec:methods-noise}

\citet{aepli-sennrich-2022-improving} introduced a simple method for improving transfer from a language to a closely related variety by inserting character-level noise into the training data.
Training on the noised data can make a model more robust to spelling variation that results in subword tokenization differences.
This method has shown to be beneficial in several studies of transfer to closely related languages and dialects \cite{aepli-sennrich-2022-improving, srivastava-chiang-2023-bertwich, srivastava-chiang-2023-fine, brahma-etal-2023-selectnoise, blaschke-etal-2023-manipulating, blaschke-etal-2024-maibaam}.

We use the machine-translated Norwegian data and randomly select a given percentage of the alphabetic\footnote{We ignore punctuation and other symbols as well as numbers written as digits.} words in a sentence.
For each of the selected words, we pick a random position within that word, and delete a character and/\allowbreak{}or insert one of the alphabetic characters that appear in the Norwegian development set.
We implement this once for each of the three PLMs, and compare selecting 10, 20, and 30\% of the words.

\paragraph{Shared task submission}
We submit the mDeBERTa model trained on data with 20\% noised words as an intent classification model.\footnote{\texttt{mainlp\_intents3\_mdeberta\_sidnor20\_5678}} 

\subsection{Auxiliary Tasks}
\label{sec:methods-auxtasks}

In another set of experiments, we include auxiliary tasks to potentially teach the model tasks related to slot filling and/or relevant information about Norwegian (or Norwegian dialects).
Previous studies on training SID models on auxiliary tasks have found that these tasks have different effects on intent detection and slot filling \cite{van-der-goot-etal-2021-masked, krueckl2025improving}.

Since we are especially interested in whether training on Norwegian auxiliary data can add useful language information to the cross-lingually evaluated English SID model, we use ScandiBERT due to its strong baseline performance when trained on the English SID data. 
For comparison, we repeat the experiments with the machine-translated Norwegian SID data.

In each of our auxiliary task experiments, we add one additional task to SID. 
The model parameters are shared across tasks, except for the task-specific decoders.
We compare two set-ups: \textbf{joint multi-task learning} (where the model is simultaneously learning SID and the other task, cf.\ \citealp{ruder2017overviewmultitasklearningdeep}), and \textbf{intermediate-task training} (where the model is first trained on the auxiliary task, and afterwards on the SID data, cf.\ \citealp{pruksachatkun-etal-2020-intermediate}).

Prior work suggests that choosing auxiliary tasks that are similar to the target task is beneficial for both multi-task learning \cite{schroder-biemann-2020-estimating} and intermediate-task training \cite{poth-etal-2021-pre, padmakumar-etal-2022-exploring}.
Training on target-language tasks in cross-lingual set-ups has yielded mixed results \cite{van-der-goot-etal-2021-masked, montariol-etal-2022-multilingual, krueckl2025improving}.
We include the following auxiliary tasks, which are either in the target dialects or similar to slot filling:

\paragraph{Dialect identification}
We use NoMusic's development data for dialect classification (with the same 90:10 split as in~\S\ref{sec:methods-baselines}) and classify instances on the dialect group level (North Norwegian, Trønder, West Norwegian, or Bokmål).

\paragraph{Part-of-speech tagging and dependency parsing}
To potentially teach the models about Norwegian sentence structure, we use the part-of-speech (POS) and syntactic dependency annotations of the UD Nynorsk LIA \cite{ovrelid-etal-2018-lia} treebank.
The dataset contains transcriptions of dialectological interviews.\footnote{The dialect distribution of this treebank is different than that of NoMusic, with around 30\% each of East, North, and West Norwegian sentences, and 7\% Trønder.
We use the script by \citet{blaschke-etal-2023-manipulating} to merge the phonetic transcriptions with the treebank.}
We use LIA's phonetic transcriptions and adjust the spelling to be somewhat more natural (Appendix~\S\ref{sec:appendix-spelling}).
We only include transcribed dialectal material (i.e., exclude utterances by interviewers), leaving 2.3k training and 622 development sentences.
We treat POS tagging and dependency parsing as two separate auxiliary tasks.

We note that some of the treebank's dependency annotations violate the Universal Dependencies \cite{de-marneffe-etal-2021-universal} standards and the treebank has been retired from official releases.
Nevertheless, we believe that it contains valuable information about Norwegian sentence structure.

\paragraph{Named entity recognition (NER)}
NER has been useful in other multi-task SID work \cite{krueckl2025improving}, and gold-standard named entity information has been found to boost slot-filling performance \citep{yao13b_interspeech}.
As no dialectal NER datasets are available, we use the NorNE dataset \cite{jorgensen-etal-2020-norne} with a reduced label set (person, organization, location, product, event, derived words).\footnote{We merge the two geo-political entity types \texttt{GPE\_LOC} and \texttt{GPE\_ORG} into location and organization, respectively, and entirely remove the category of miscellaneous entities, since it occurs very rarely in the dataset.}
The dataset contains 29.9k training and 4.3k development set sentences 
(slightly more than half are in Bokmål, and the rest in the other written standard, Nynorsk).

\paragraph{Shared task submission}
We submit the model first trained on the dependency data and subsequently on xSID's English data for slot filling.\footnote{\texttt{mainlp\_slots3\_scandibert\_deprel\_sid\_8446}}

\subsection{Layer Swapping}
\label{sec:methods-layer-swapping}
Layer Swapping was recently proposed as a method for cross-lingual transfer \cite{bandarkar2024layerswappingzeroshotcrosslingual}. 
The authors fine-tune a task expert on English instruction data, and a language expert on general-purpose data in the target language. 
They replace the top and bottom layers of the task expert with the corresponding layers of the language expert, producing a model capable of performing the task in the target language. 
We adapt this method -- originally applied to \textsc{Llama~3.1~8B} \cite{grattafiori2024llama3herdmodels}, a 32-layer decoder model -- to a 12-layer encoder model.

\paragraph{Experts}
We use mDeBERTa \cite{he2023debertav3}, as it is the strongest baseline when fine-tuned on the NoMusic training data. We replace layers of an \textbf{EnSID expert} with layers from a \textbf{Norwegian expert}, and consider different options for the latter.

To produce the EnSID expert, we jointly fine-tune on the English xSID training data for both slot filling and intent classification. We build on a JointBERT implementation (closely following \citealp{chen2019bertjointintentclassification}),\footnote{\href{https://github.com/monologg/JointBERT}{\texttt{https://github.com/monologg/JointBERT}} (\href{https://github.com/monologg/JointBERT?tab=Apache-2.0-1-ov-file\#readme}{Apache~2.0}; commit \texttt{
00324f6})} using default hyperparameters, which do not include a CRF and specify 10 epochs. The best checkpoint is chosen based on performance on the NoMusic development set.

We consider four options for the Norwegian expert: the NoMusic dialect baseline described in~\S\ref{sec:methods-baselines} (here referred to as the \textbf{NorSID expert}), as well as three Norwegian language experts.

To produce the language experts, we fine-tune\footnote{mDeBERTa is trained using replaced token detection (RTD; \citealp{he2023debertav3}) rather than MLM, hence we do not consider MLM fine-tuning a continuation of training.} with the masked language modeling (MLM) objective using an example from Sentence Transformers \cite{reimers-gurevych-2019-sentence}.\footnote{\href{https://github.com/UKPLab/sentence-transformers/blob/master/examples/unsupervised_learning/MLM/train_mlm.py}{\texttt{https://github.com/UKPLab/sentence-\allowbreak{}transformers} \href{https://github.com/UKPLab/sentence-transformers?tab=Apache-2.0-1-ov-file}{(Apache~2.0; commit \texttt
{1cb196a})}}} We train for 20 epochs and select the best checkpoint based on perplexity on the development set of NoMusic.

We use three different datasets for the language experts to examine whether the text style/\allowbreak{}genre and language variety makes a difference: the Bokmål transcriptions of interviews in the Nordic Dialect Corpus (NDC; \citealp{johannessen-etal-2009-nordic}),\footnote{We use a random 80:10:10 split of half of the corpus.} the Bokmål part of the Norwegian Dependency Treebank (NDT; \citealp{solberg-etal-2014-norwegian}) which contains news articles, blog posts, and government reports/\allowbreak{}transcripts,\footnote{This treebank is the basis of the NorNE dataset~(\S\ref{sec:methods-auxtasks}).} 
and the NoMusic training set.

\paragraph{Identifying layers to replace}
As an ablation experiment to identify layers of the EnSID expert that might be replaceable, we revert its layers back to their state in the pretrained model and observe the performance of the resulting model on the NoMusic development set. For each of the mDeBERTa models fine-tuned on the English xSID data with three different seeds, we revert pairs of sequential layers (i.e., 0,1, then 1,2, and so on).

Unlike \citet{bandarkar2024layerswappingzeroshotcrosslingual}, we are unable to use the mean absolute value (MAV) of the difference in parameters through fine-tuning to identify less salient layers. The variance in change of the parameters of the EnSID expert is very small at $1.5\times10^{-7}$, such that no layers exhibit significantly higher MAVs than others. This may be due to any number of differences of our setup, such as model architecture, layer depth, fine-tuning objective, amount of fine-tuning data, or simply duration of fine-tuning; further analysis of layer-wise training dynamics is left to future work.

\paragraph{Model assembly}
The layer-reverting experiments identify the first two layers of the EnSID expert as suited for replacement. We replace the token embeddings and the first two encoder layers of the EnSID expert with the corresponding layers of the Norwegian expert, resulting in four assembled models, one for each Norwegian expert. 
We do not merge any parameters.

\paragraph{Shared task submission}
We submit the model produced by assembling layers of the NorSID expert and the EnSID expert.\footnote{\texttt{mainlp\_\{slots,intents\}2\_mdeberta\_topline\_\allowbreak{}swapped}}

\section{Results and Analysis}
\label{sec:results}

In this section, we mainly focus on the test scores.
For the shared task, we submitted models considering their development set performance.
These are denoted by asterisks in the results tables, and further discussed in~\S\ref{sec:results-shared-task}.
All models were trained (and evaluated on the development set) before the test set was released.

Table~\ref{tab:results-all} in Appendix~\ref{sec:appendix-results} shows the development and test scores for all systems.

\subsection{Baselines}
\label{sec:results-baselines}

The training data choice had a greater effect on the SID quality than the PLM choice (Table~\ref{tab:results-baselines}).

\begin{table}[t]
\setlength{\tabcolsep}{3pt}
\centering
\adjustbox{max width=\linewidth}{%
\begin{tabular}{@{}llrrr@{\,\,}r@{}}
\toprule
\textbf{Training data} & \textbf{Model} & \multicolumn{1}{l}{\textbf{Intents}} && \multicolumn{1}{l}{\textbf{Slots}} \\ \midrule
English & NorBERT & \cellcolor[HTML]{95D4B5}95.1\,\textsubscript{0.2} && \cellcolor[HTML]{B3E1CA}79.7\,\textsubscript{0.4} \\
(train) & ScandiBERT~ & \cellcolor[HTML]{A0D9BD}94.8\,\textsubscript{0.8} && \cellcolor[HTML]{99D6B8}80.7\,\textsubscript{0.7} \\
 & mDeBERTa & \cellcolor[HTML]{FFFFFF}92.4\,\textsubscript{1.8} && \cellcolor[HTML]{FEFEFE}76.5\,\textsubscript{1.2} \\
\horizontalspacer
Norwegian & NorBERT & \cellcolor[HTML]{6CC499}96.2\,\textsubscript{0.5} && \cellcolor[HTML]{CCCCCC}53.9\,\textsubscript{0.3} \\
(train, MT) & ScandiBERT & \cellcolor[HTML]{67C295}96.3\,\textsubscript{0.1} && \cellcolor[HTML]{CDCDCD}54.6\,\textsubscript{0.4} \\
 & mDeBERTa & \cellcolor[HTML]{57BB8A}96.7\,\textsubscript{0.3} && \cellcolor[HTML]{CECECE}55.2\,\textsubscript{1.1} \\
\horizontalspacer
Nor. dialect & NorBERT & \cellcolor[HTML]{B8E2CE}94.2\,\textsubscript{0.6} && \cellcolor[HTML]{FFFFFF}76.8\,\textsubscript{1.1} \\
(dev, 90\%) & ScandiBERT & \cellcolor[HTML]{EFF9F4}92.8\,\textsubscript{0.6} && \cellcolor[HTML]{8DD1AF}81.2\,\textsubscript{0.6} \\
 & mDeBERTa & \cellcolor[HTML]{D6EFE3}93.4\,\textsubscript{0.7} && \cellcolor[HTML]{57BB8A}83.2\,\textsubscript{1.0} & *\\ \bottomrule
\end{tabular}
}
\caption{\textbf{Test scores of baseline models} (intent accuracy in~\%, slot span \fonescore{} in~\%) trained on English data, machine-translated Norwegian data, or 90\% of the dialectal Norwegian development set.
The results are averaged over three runs, with standard deviations as subscripts.
*\,Model submitted to the shared task (slots and intents).}
\label{tab:results-baselines}
\end{table}

\paragraph{Training data}
Despite the language difference, the models trained on the English training data provide strong baselines -- especially for the Norwegian and Scandinavian PLMs, which achieve intent prediction accuracies of 94.8--95.1\% and slot-filling \fonescore{} scores of 79.7--80.7\%.

The models trained on the machine-translated Norwegian training set produce better intent labels (with accuracies between 96.2 and 94.8\%), but are poor slot fillers (53.9--55.2\% \fonescore{}).\footnote{%
This is similar to the results by \cite{van-der-goot-etal-2021-masked}, who find training on translated data to be beneficial for intent classification. 
In their experiments, translated data improves slot filling for a PLM with poor baseline scores for cross-lingual slot filling, but lowers the performance of another model whose cross-lingual slot-filling scores were already quite high when trained on English data.
}
We suspect this is due to quality issues with the slot label projections.
To substantiate this, we compare the strict span~\fonescore{} scores with their loose counterpart, which allows spans that only partially overlap.
Although the models trained on machine-translated Norwegian achieve much lower strict~\fonescore{} scores, the loose~\fonescore{} scores are similar to those of the other baselines (Table~\ref{tab:results-baselines-slots-detailed}, Appendix~\ref{sec:appendix-results}). 
This suggests that the slot annotations of the machine-translated Norwegian baselines mainly suffer from incorrect spans, as would be expected from poor projections, which affect the span, but not the label.

Training the models on the largely dialectal development set led to overfitting -- these models show the greatest drop between development and test set performance (Table~\ref{tab:results-all} in Appendix~\ref{sec:appendix-results}). 
This may have been exacerbated by how we stratified the data, as we did not ensure that all translations of the same sentence were assigned to the same split. 
Furthermore, the development set is significantly smaller than the training set (2.9k vs.\ 43.6k samples).
Finally, one intent and one slot type were present in the test but not in the development set, as well as seven \texttt{I} labels (though the corresponding \texttt{B} was seen, more on this under \nameref{sec:limitations}). 
Despite all of this, the models fine-tuned on this dataset produce some of the best slot annotations (with \fonescore{} scores between 76.8 and 83.2\%).

\paragraph{PLM}
No PLM is consistently the best or worst model.
For the models trained on the English or machine-translated Norwegian data, performance on slot filling appears to be correlated with performance on intent classification, and vice versa.
However, there seems to be no relation between the two for the models trained on the dialectal data where, e.g., NorBERT produces the best intent labels but the worst slot annotations.

\subsection{Character-Level Noise}
\label{sec:results-noise}
\begin{table}[t]
\adjustbox{max width=\linewidth}{%
\begin{tabular}{@{}l@{}rr@{\,}r@{\,}rr@{\,}r@{\,}r@{\,\,}r@{}}
\toprule
\multicolumn{2}{@{}r@{}}{\textbf{PLM} \hfill \textbf{Noise (\%)}} & \multicolumn{2}{c}{\textbf{Intents}} & \multicolumn{1}{c}{\boldmath$\Delta$} & \multicolumn{2}{c}{\textbf{Slots}} & \multicolumn{1}{c}{\boldmath$\Delta$} \\ \midrule
NorBERT & 0 & 96.2 & \textsubscript{0.5} &  & 53.9 & \textsubscript{0.3} &  \\
 & 10 & 96.4 & \textsubscript{0.3} & \multicolumn{1}{r}{\cellcolor[HTML]{F3FAF7}+0.2} & 55.1 & \textsubscript{0.8} & \multicolumn{1}{r}{\cellcolor[HTML]{BEE5D2}+1.2} \\
 & 20 & 97.2 & \textsubscript{0.2} & \multicolumn{1}{r}{\cellcolor[HTML]{C7E9D8}+1.0} & 55.0 & \textsubscript{0.3} & \multicolumn{1}{r}{\cellcolor[HTML]{C2E6D4}+1.1} \\
 & 30 & 97.4 & \textsubscript{0.5} & \multicolumn{1}{r}{\cellcolor[HTML]{BDE4D1}+1.2} & 54.1 & \textsubscript{1.1} & \multicolumn{1}{r}{\cellcolor[HTML]{F7FCFA}+0.1} \\
\horizontalspacer
ScandiBERT & 0 & 96.3 & \textsubscript{0.1} &  & 54.6 & \textsubscript{0.4} &  \\
 & 10 & 96.5 & \textsubscript{0.4} & \multicolumn{1}{r}{\cellcolor[HTML]{F6FCF9}+0.2} & 55.9 & \textsubscript{0.7} & \multicolumn{1}{r}{\cellcolor[HTML]{B5E1CC}+1.3} \\
 & 20 & 97.5 & \textsubscript{0.2} & \multicolumn{1}{r}{\cellcolor[HTML]{BCE4D1}+1.2} & 54.6 & \textsubscript{0.5} & \multicolumn{1}{r}{\cellcolor[HTML]{FEFDFD}--0.0} \\
 & 30 & 97.1 & \textsubscript{0.5} & \multicolumn{1}{r}{\cellcolor[HTML]{D5EEE2}+0.8} & 54.8 & \textsubscript{0.5} & \multicolumn{1}{r}{\cellcolor[HTML]{F6FBF9}+0.2} \\
\horizontalspacer
mDeBERTa & 0 & 96.7 & \textsubscript{0.3} &  & 55.2 & \textsubscript{1.1} &  \\
 & 10 & 96.5 & \textsubscript{0.9} & \multicolumn{1}{r}{\cellcolor[HTML]{FDF4F4}--0.2} & 55.6 & \textsubscript{1.0} & \multicolumn{1}{r}{\cellcolor[HTML]{EAF7F1}+0.4} \\
 & 20 & 97.5 & \textsubscript{0.2} & \multicolumn{1}{r}{\cellcolor[HTML]{D4EEE1}+0.8} & 55.5 & \textsubscript{0.5} & \multicolumn{1}{r}{\cellcolor[HTML]{F2FAF6}+0.2} & * \\
 & 30 & 97.0 & \textsubscript{0.5} & \multicolumn{1}{r}{\cellcolor[HTML]{EDF8F3}+0.3} & 56.2 & \textsubscript{0.5} & \multicolumn{1}{r}{\cellcolor[HTML]{C7E9D8}+1.0} \\ \bottomrule
\end{tabular}
}
\caption{\textbf{Test scores of models trained on noised data} (intent accuracy in~\%, slot span \fonescore{} in~\%).
The results are averaged over three runs, with standard deviations as subscripts. 
The $\Delta$ columns show the differences to the respective baseline (0\,\% noise).
*\,Model submitted to the shared task (intents).
}
\label{tab:results-noise}
\end{table}

Fine-tuning on noised data generally improves the models' performance (Table~\ref{tab:results-noise}) -- by up to 1.2~percentage points (pp.) for intent classification and up to 1.3~pp.\ for slot filling.
Which noise level helps most depends on the PLM choice; this is similar to previous findings on using noised data for POS tagging in Norwegian dialects and other language varieties \cite{blaschke-etal-2023-manipulating}.
However, the effect of noise also depends on the task -- the trends are different for intent classification and slot filling.

Prior work has found the ratios of words that were split into multiple subword tokens to be a strong predictor for transfer success between closely related varieties: the more similar the split word ratios are in the training and evaluation data, the more successful transfer tends to be \cite{blaschke-etal-2023-manipulating}.
In our study, only the intent classification results correlate with this difference in split word ratio (Table~\ref{tab:noise-correlations} in Appendix~\ref{sec:appendix-results}).
We hypothesize that the weak correlations with the slot-filling results might be due to the mixed quality of the silver-standard slot annotations in the training data.

\subsection{Auxiliary Tasks}
\label{sec:results-auxtasks}
\begin{table}[t]
\centering
\adjustbox{max width=\linewidth}{%
\begin{tabular}{@{}l@{\hspace{4pt}}lr@{\,}rrr@{\,}rr@{}r@{}}
\toprule
\multicolumn{2}{@{}l}{\textbf{Task}} & \multicolumn{2}{c}{\textbf{Intents}} & \multicolumn{1}{c}{\boldmath$\Delta$} & \multicolumn{2}{c}{\textbf{Slots}} & \multicolumn{1}{c}{\boldmath$\Delta$} \\ \midrule
\multicolumn{9}{l}{\textit{English training data}} \\
\multicolumn{2}{@{}l}{Baseline} & 94.8 & \textsubscript{0.8} &  & 80.7 & \textsubscript{0.7} &  \\
\horizontalspacer
Dial&\multitask{} & 83.8 & \textsubscript{3.2} & \cellcolor[HTML]{E67C73}--11.0 & 75.8 & \textsubscript{1.2} & \cellcolor[HTML]{F3C4C0}--4.9 \\
&\sequential{} & 94.0 & \textsubscript{1.7} & \cellcolor[HTML]{FDF4F4}--0.8 & 79.2 & \textsubscript{0.9} & \cellcolor[HTML]{FBEDEB}--1.5 \\
\horizontalspacer
POS&\multitask{} & 94.9 & \textsubscript{0.3} & \cellcolor[HTML]{FFFFFF}+0.0 & 81.1 & \textsubscript{0.3} & \cellcolor[HTML]{F1F9F5}+0.4 \\
&\sequential{} & 94.7 & \textsubscript{0.2} & \cellcolor[HTML]{FEFDFC}--0.2 & 82.2 & \textsubscript{1.1} & \cellcolor[HTML]{CFECDE}+1.5 \\
\horizontalspacer
Dep&\multitask{} & 93.5 & \textsubscript{0.6} & \cellcolor[HTML]{FCEFEE}--1.3 & 81.5 & \textsubscript{0.2} & \cellcolor[HTML]{E6F5ED}+0.8 \\
&\sequential{} & 94.9 & \textsubscript{1.2} & \cellcolor[HTML]{FEFFFE}+0.1 & 81.8 & \textsubscript{0.7} & \cellcolor[HTML]{DBF1E6}+1.1 & *\\
\horizontalspacer
NER&\multitask{} & 95.3 & \textsubscript{1.0} & \cellcolor[HTML]{EFF9F4}+0.5 & 80.6 & \textsubscript{1.0} & \cellcolor[HTML]{FEFDFD}--0.1 \\
&\sequential{} & 95.0 & \textsubscript{0.4} & \cellcolor[HTML]{FBFEFD}+0.1 & 81.1 & \textsubscript{0.9} & \cellcolor[HTML]{F1FAF6}+0.4 \\
\horizontalspacer
\midrule
\multicolumn{9}{l}{\textit{Machine-translated Norwegian training data}} \\
\multicolumn{2}{@{}l}{Baseline} & 96.3 & \textsubscript{0.1} &  & 54.6 & \textsubscript{0.4} &  \\
\horizontalspacer
Dial&\multitask{} & 89.2 & \textsubscript{1.4} & \cellcolor[HTML]{EEAAA4}--7.1 & 51.7 & \textsubscript{0.1} & \cellcolor[HTML]{F8DBD9}--2.9 \\
&\sequential{} & 95.2 & \textsubscript{1.0} & \cellcolor[HTML]{FCF2F1}--1.1 & 53.7 & \textsubscript{0.5} & \cellcolor[HTML]{FCF4F3}--0.9 \\
\horizontalspacer
POS&\multitask{} & 96.8 & \textsubscript{0.4} & \cellcolor[HTML]{F1FAF5}+0.4 & 53.7 & \textsubscript{0.6} & \cellcolor[HTML]{FCF4F3}--0.9 \\
&\sequential{} & 96.7 & \textsubscript{0.4} & \cellcolor[HTML]{F4FBF7}+0.3 & 54.4 & \textsubscript{0.8} & \cellcolor[HTML]{FEFCFC}--0.2 \\
\horizontalspacer
Dep&\multitask{} & 96.9 & \textsubscript{0.3} & \cellcolor[HTML]{EDF8F3}+0.5 & 53.7 & \textsubscript{0.2} & \cellcolor[HTML]{FCF4F3}--0.9 \\
&\sequential{} & 96.4 & \textsubscript{0.3} & \cellcolor[HTML]{FBFEFC}+0.1 & 54.8 & \textsubscript{0.6} & \cellcolor[HTML]{F9FDFB}+0.2 \\
\horizontalspacer
NER&\multitask{} & 96.9 & \textsubscript{0.1} & \cellcolor[HTML]{ECF8F2}+0.6 & 53.8 & \textsubscript{0.3} & \cellcolor[HTML]{FDF5F5}--0.8 \\
&\sequential{} & 96.4 & \textsubscript{0.5} & \cellcolor[HTML]{FDFFFE}+0.1 & 53.5 & \textsubscript{1.0} & \cellcolor[HTML]{FCF2F1}--1.1 \\ \bottomrule
\end{tabular}
}
\caption{\textbf{Test scores of models trained on auxiliary tasks} (intent accuracy in~\%, slot span \fonescore{} in~\%).
The results are averaged over three runs, with standard deviations as subscripts. 
The $\Delta$ columns show the differences to the respective baseline.
Key: 
\textit{Dial}\,=\,dialect identification, 
\textit{dep}\,=\,dependency parsing,
\multitask{}\,=\,multitask learning,
\sequential{}\,=\,intermediate-task training.
*\,Model submitted to the shared task (slots).
}
\label{tab:results-auxtasks}
\end{table}

The effect of the auxiliary tasks depends on the tasks themselves, the language of the SID data, and whether they are trained before or simultaneously with the target SID task.
Table~\ref{tab:results-auxtasks} shows the results on the SID test data; Table~\ref{tab:auxtask-details} in Appendix~\ref{sec:appendix-results} also shows the development scores on the SID and auxiliary task data.

\paragraph{Intermediate-task training vs.\ multi-task learning}
For slot filling, intermediate-task training (training first on the auxiliary task and afterwards on the SID data) generally achieves better results than simultaneous multi-task learning. 
For intent classification, there is no clear trend.

We additionally examine whether the effects of multi-task learning are similar across tasks by inspecting the models' performances on the development sets of the auxiliary tasks (Table~\ref{tab:auxtask-details} in Appendix~\ref{sec:appendix-results}).
For the auxiliary tasks, multi-task learning nearly always yields worse results than exclusively training on the auxiliary tasks (as the first step in intermediary-task training).
Although the performance gap between the two settings is especially large for the two syntactic tasks (with multi-task learning achieving scores that are 11.8--26.7~pp.\ lower), the impact on the corresponding SID performance is less clear-cut (with multi-task learning leading by up to 0.5~pp.\ in some constellations and falling behind by 2.1~pp.\ in others).

\paragraph{Auxiliary task choice and SID training language}
Dialect identification diminishes both the intent classification and slot-filling performance in all of our set-ups (most drastically in the multi-task set-up with the English SID data, with drops of 11.0~pp.\ for intent classification and 4.9~pp.\ for slot filling).

The effects of the other tasks depend on the SID training language.
For the models fine-tuned on Norwegian data, the other tasks slightly improve intent classification performance (with gains of up to 0.6~pp.)\ but typically negatively impact slot filling (with changes between +0.2 and --0.9~pp.) -- the grammatical tasks do not mitigate the effect of poor slot annotations in the machine-translated data.

For the English SID training data, the syntax-related tasks (POS tagging and dependency parsing) improve slot filling by between 0.4 and 1.5~pp., but have no or a negative effect on the intent classification performance (changes to the baseline between +0.1 and --1.3~pp.).
Despite positive prior findings \cite{krueckl2025improving},
NER has no or only slightly positive effects on either SID task.

\paragraph{Dialects}
There is no apparent connection between the dialect distributions in the auxiliary task training data and the SID performance on the different dialect groups (Table~\ref{tab:auxtasks-dialects} in Appendix~\ref{sec:appendix-results}).
This applies both to the models trained on English SID data and on the Norwegian translations, although the gains per dialect group differ between them.

For the syntactic tasks, one possible explanation is that the dialect transcriptions do not sufficiently align with the ad-hoc dialect spellings used in NoMusic to show strong effects based on the represented dialect groups.

\subsection{Layer Swapping}
\label{sec:results-layer-swapping}
\paragraph{Identifying layers to replace}
Results of reverting pairs of layers of the EnSID expert are shown in Table~\ref{tab:results-layer-reverting}. We found that in general, performance decreased as later layers were reverted. This aligns with our intuition that the later layers, being closer to the classification heads, are particularly important for performance.

Notably, we found that reverting layers 0 and 1 slightly increased performance on both slot filling and intent classification (across three runs we observed an average improvement of slot \fonescore{} of 3.0~pp.\ and intent accuracy of 0.9~pp.). This improvement through reverting is somewhat surprising, and suggests that something about the fine-tuning process on the English data is counterproductive to the robustness of the model to out-of-language data, at least where the first two layers are concerned.

We also observed a large variance in the effect of reverting the last two layers on intent classification, this is due to the first seed seeing quite a large drop (to 55.7\%, the average accuracy of the other two seeds was 86.1\%).
\begin{table}
\centering
\adjustbox{max width=\linewidth}{%
\begin{tabular}{@{}lr@{\,}rrr@{\,}rr}
\toprule
\textbf{Layers} & \multicolumn{2}{l}{\textbf{Intents}} & \multicolumn{1}{c}{\textbf{}$\Delta$}     & \multicolumn{2}{l}{\textbf{Slots}} & \multicolumn{1}{c}{\textbf{}$\Delta$}     \\
\midrule
\textit{none}            & 95.1         & \textsubscript{0.6} & \multicolumn{1}{l}{}          & 77.1         & \textsubscript{1.4}        & \multicolumn{1}{l}{}          \\
0,1             & 96.1         & \textsubscript{0.4}          & \cellcolor[HTML]{D8EFE4}+0.9   & 80.1         & \textsubscript{0.6}        & \cellcolor[HTML]{82CDA8}+3.0   \\
1,2             & 96.0         & \textsubscript{0.2}          & \cellcolor[HTML]{DBF1E6}+0.9   & 77.8         & \textsubscript{1.0}        & \cellcolor[HTML]{E5F5ED}+0.6   \\
2,3             & 95.1         & \textsubscript{0.8}          & \cellcolor[HTML]{FFFFFF}0.0   & 69.9         & \textsubscript{0.5}        & \cellcolor[HTML]{F7D9D6}--7.2  \\
3,4             & 93.9         & \textsubscript{0.6}          & \cellcolor[HTML]{FDF8F8}--1.2  & 65.8         & \textsubscript{0.2}        & \cellcolor[HTML]{F3C3BF}--11.3 \\
4,5             & 93.8         & \textsubscript{0.7}          & \cellcolor[HTML]{FDF7F7}--1.4  & 68.0         & \textsubscript{1.6}        & \cellcolor[HTML]{F5CFCB}--9.1  \\
5,6             & 93.6         & \textsubscript{1.1}          & \cellcolor[HTML]{FDF7F6}--1.5  & 69.3         & \textsubscript{2.0}        & \cellcolor[HTML]{F7D6D3}--7.8  \\
6,7             & 90.7         & \textsubscript{0.9}          & \cellcolor[HTML]{FAE7E6}--4.5  & 63.1         & \textsubscript{5.8}        & \cellcolor[HTML]{F0B5B0}--14.0 \\
7,8             & 87.0         & \textsubscript{1.2}          & \cellcolor[HTML]{F6D4D1}--8.1  & 59.7         & \textsubscript{3.8}        & \cellcolor[HTML]{EDA39D}--17.4 \\
8,9             & 84.3         & \textsubscript{2.7}          & \cellcolor[HTML]{F4C6C2}--10.8 & 58.0         & \textsubscript{2.5}        & \cellcolor[HTML]{EB9A94}--19.1 \\
9,10            & 72.6         & \textsubscript{9.4}          & \cellcolor[HTML]{E88880}--22.6 & 54.1         & \textsubscript{5.5}        & \cellcolor[HTML]{E7867D}--23.0 \\
10,11           & 76.0         & \textsubscript{17.9}        & \cellcolor[HTML]{EB9A93}--19.2 & 59.4         & \textsubscript{0.6}        & \cellcolor[HTML]{EDA29B}--17.7\\
\bottomrule
\end{tabular}
}
\caption{\textbf{Development scores of the EnSID expert with reverted layers} (intent accuracy in~\%, slot span \fonescore{} in~\%).
The results are averaged over three runs with standard deviations as subscripts.}
\label{tab:results-layer-reverting}
\end{table}

\paragraph{Choosing a complementary expert}
Table~\ref{tab:results-layer-swapping-other-models} shows the results of replacing the first two layers of the EnSID expert with the corresponding layers of each of our Norwegian experts. These combinations performed roughly on par with or slightly better than the reverted model, except for the model containing the layers from the NorSID expert, which performed better, particularly for slot filling. Further analysis is needed to better understand what makes layers useful for assembling into a model, this is left for future work.
\begin{table}[t]
\adjustbox{max width=\linewidth}{%
\begin{tabular}{@{}lllll}
\toprule
\textbf{Norwegian Expert}& {\textbf{Intents}}  & \multicolumn{1}{c}{\boldmath$\Delta$} & {\textbf{Slots}}  &   \multicolumn{1}{c}{\boldmath$\Delta$}   \\
\midrule
N/A -- EnSID unchanged  & {95.1} {\textsubscript{0.6}} & & {77.1} {\textsubscript{1.4}} & \\
N/A -- EnSID reverted (0,1) & {96.1} {\textsubscript{0.4}} & {\cellcolor[HTML]{D8EFE4}+0.9} & {80.1} {\textsubscript{0.6}} & {\cellcolor[HTML]{E6F5EE}+3.0} \\
NorSID expert & {98.3} {\textsubscript{0.4}} & {\cellcolor[HTML]{83CDA9}+2.2} & {86.5} {\textsubscript{0.6}} & {\cellcolor[HTML]{AFDFC7}+9.6} \\
NoMusic MLM & {96.9} & {\cellcolor[HTML]{D2EDE0}+0.8} & {78.8} & {\cellcolor[HTML]{F1FAF6}+1.7} \\
NDT MLM   & {97.4} & {\cellcolor[HTML]{B6E2CC}+1.3} & {78.6} &  {\cellcolor[HTML]{F3FAF7}+1.5} \\
NDC MLM   & {96.3} & {\cellcolor[HTML]{F3FBF7}+0.2} & {77.9} &  {\cellcolor[HTML]{F9FDFB}+0.8} \\
\bottomrule
\end{tabular}
}
\caption{\textbf{Development scores of \textit{assembled} models using different Norwegian experts} (intent accuracy in~\%, slot span \fonescore{} in~\%). Each Norwegian expert is assembled with the EnSID expert. Results for the unchanged EnSID expert and the best reverted model, layers 0,1, are shown for comparison, each averaged over three runs. The assembled model with the NorSID expert is averaged over nine runs (for each combination of NorSID and EnSID expert). We don't repeat runs for unpromising language experts. The standard deviation, where applicable, is denoted by subscripts.}
\label{tab:results-layer-swapping-other-models}
\end{table}

As these were exploratory preliminary experiments, we do not repeat runs for unpromising language experts.

\paragraph{Final submission}
Results for the submitted assembled model (layers from the EnSID and NorSID expert), and the individual experts on both NoMusic and the xSID~0.6 English set are shown in Table~\ref{tab:results-layer-swapping}. Overall, the assembled model is more robust to out-of-language data than the respective experts, outperforming the EnSID expert on the Norwegian development and test sets, and mostly outperforming the Norwegian expert on the English sets, except for intent classification on the test set. We hypothesize that this exception may be due to the EnSID expert overfitting the intent classification task, which was not mitigated by using the first two layers of the Norwegian SID expert.

Using only two layers of the Norwegian SID expert, which suffered from overfitting (\S\ref{sec:results-baselines}), seems to have a regularizing effect, as the assembled model outperforms the Norwegian SID expert in both tasks on the Norwegian test set.
\begin{table}
\adjustbox{max width=\linewidth}{%
\begin{tabular}{@{}l@{\,}r@{\,}r@{\,}r@{\,}rr@{\,}rr@{\,}r@{}}
\toprule
& \multicolumn{4}{c}{\textbf{Intents}}             & \multicolumn{4}{c}{\textbf{Slots}}     \\
\cmidrule(lr){2-5} \cmidrule(lr){6-9}
\multicolumn{3}{r}{\textbf{dev (no)}} & \multicolumn{2}{@{}r}{\textbf{test (no)}} & \multicolumn{2}{@{}r}{\textbf{dev (no)}} & \multicolumn{2}{@{}r@{}}{\textbf{test (no)}} \\
\midrule
EnSID expert  & 95.1       & \textsubscript{0.6}     & 92.0       & \textsubscript{0.8}      & 78.6      & \textsubscript{1.1}      & 77.2       & \textsubscript{1.6}      \\
NorSID expert & \textbf{99.4}       & \textsubscript{0.0}     & 93.4       & \textsubscript{0.7}      & \textbf{96.4}      & \textsubscript{0.4}      & 83.2       & \textsubscript{1.0}      \\
Assembled*        & 98.3       & \textsubscript{0.4}     & \textbf{96.4}       & \textsubscript{0.2}      & 86.5      & \textsubscript{0.6}      & \textbf{84.9}       & \textsubscript{0.5}      \\
\midrule
\multicolumn{3}{r}{\textbf{dev (en)}}  & \multicolumn{2}{@{}r}{\textbf{test (en)}}  & \multicolumn{2}{@{}r}{\textbf{dev (en)}}  & \multicolumn{2}{@{}c@{}}{\textbf{test (en)}}  \\
\midrule
EnSID expert  & \textbf{100.0}      & \textsubscript{0.0}     & 99.2       & \textsubscript{0.0}      & 97.1      & \textsubscript{0.3}      & \textbf{96.0}       & \textsubscript{0.3}      \\
NorSID expert & \textbf{100.0}      & \textsubscript{0.0}     & \textbf{100.0}      & \textsubscript{0.0}      & 90.1      & \textsubscript{1.0}      & 80.9       & \textsubscript{1.4}      \\
Assembled*        & \textbf{100.0}      & \textsubscript{0.0}     & 99.3       & \textsubscript{0.2}      & \textbf{97.5}      & \textsubscript{0.3}      & \textbf{96.0}       & \textsubscript{0.3}\\
\bottomrule
\end{tabular}
}
\caption{\textbf{Development and test scores of the original experts and assembled model on NoMusic (no) and xSID 0.6 English (en)} (intent accuracy and slot \fonescore{} in~\%, best results bolded).
The results are averaged over three runs for the experts, and over nine runs for the assembled model, with standard deviations as subscripts.
*\,Model submitted to the shared task (slots and intents).}
\label{tab:results-layer-swapping}
\end{table}

\subsection{Results of Shared Task Submissions}
\label{sec:results-shared-task}
We submitted three systems per task (slot and intent detection) and did not participate in dialect classification. 
The official results are provided in the shared task overview paper \cite{scherrer-etal-2025-vardial} 
and the accompanying website,\footnote{\url{https://github.com/ltgoslo/NoMusic/blob/main/NorSID/results.md}} and we include them in Table~\ref{tab:results-all} (Appendix~\ref{sec:appendix-results}).
Unlike the previous sections, they only represent a single random seed.
Of our intent classification systems, noise injection worked best (ranked 5th of all submissions; 97.64\% accuracy), narrowly followed by Layer Swapping (6th rank; 97.16\%).
Both beat the baseline trained only on the dialectal development set (10th rank; 93.47\%).

For slot detection, Layer Swapping instead was our best method, ranking third in the competition (85.57\%~\fonescore).
Compared to our other two submissions -- the baseline trained on the development set (5th rank; 83.68\%) and the model with intermediate-task training on dependency parsing (6th rank; 82.57\%) -- it performed best on three out of the four Norwegian varieties.

\section{Discussion and Conclusion}
\label{sec:discussion}
The strength of our baselines suggest that the NorSID task is, relatively speaking, less challenging than other dialectical variants of xSID (cf.\ \citealp{van-der-goot-etal-2021-masked, aepli-etal-2023-findings, srivastava-chiang-2023-fine, kwon-etal-2023-sidlr, winkler-etal-2024-slot, munoz-ortiz-etal-2025-evaluating, krueckl2025improving}). 
We suspect that there is less deviation from standard Norwegian, and less variation between the dialects. 
This limits the gains we could expect from additional methods, particularly on the intent classification task, where the accuracy of our baselines ranges from 92.4\% to 96.7\% on the test set.

We observe somewhat of a trade-off between performance on intent classification (strongest for models trained on Norwegian data) and slot filling (strongest for models trained on the gold-standard English training or Norwegian development data; \S\ref{sec:results-baselines}).
We hypothesize that the latter is due to the poor quality of the slot labels in the machine-translated Norwegian training data.

We see noise injection as a simple way to improve transfer between a standard language and related varieties~(\S\ref{sec:results-noise}), although it requires access to appropriate training data.
Where a language has enough resources for additional annotated datasets, we see mixed effects from the inclusion of auxiliary NLP tasks~(\S\ref{sec:results-auxtasks}).
Which auxiliary tasks help SID performance depends on the target-task training data and SID subtask (intent classification vs.\ slot filling) and remains hard to predict, requiring further research.

Improving performance on the slot-filling task proved to be quite difficult; our most successful method by a small margin is the assembled model made up of layers from a model trained on the NoMusic development set (NorSID expert), and another on the English xSID data~(\S\ref{sec:results-layer-swapping}). 
Using layers from both of these models seems to have a regularizing effect and produces a model that is able to perform well on both languages and suffers less from overfitting than the NorSID expert.

We successfully adapted Layer Swapping -- originally applied to a 32-layer decoder -- to a 12-layer encoder, demonstrating its potential for resource-efficient cross-lingual transfer. Layer Swapping could prove useful for modular solutions, as layers for different languages could dynamically replace those of a ``base'' SID expert to adapt the model. We again note that the subset of the development set of NoMusic we used, at 2.9k examples, is much smaller than the set used to train our EnSID expert, at 43.6k examples; this modular approach would allow adaptation to different languages in a fairly lightweight manner post-hoc.

We encourage further research comparing (and combining) different methods for low-resource NLP with the same training and/or evaluation data.

\section*{Limitations}
\label{sec:limitations}
Both MaChAmp and the JointBERT implementation only consider the exact labels seen during training; consequently our SID models will not predict unseen \texttt{I}~tags, even if the corresponding \texttt{B}~tag is known.
In particular, the English xSID sets have fewer \texttt{I}~tags, i.e., corresponding slots are sometimes spread over more words in NoMusic. 
We also find that the NoMusic test set has more \texttt{I}~tags than the development set.

While we compare several different approaches for improving SID on this task, we find the conditions of their success are difficult to generalize. For example, no auxiliary task has prevailed. For Layer Swapping, it is not clear what makes layers of particular expert suitable for assembly, and whether our findings generalize to other models, languages, or tasks. Further work is needed to understand which method will work best for what conditions, and how best to apply each method.

Because of time constraints, we were not able to further investigate the effect of including auxiliary datasets in standard vs.\ dialectal varieties.
In particular, it would be interesting to include POS tagging and dependency parsing on Bokmål or Nynorsk data (e.g., the NDT and LIA treebanks we used in other ways in this paper).

Similarly, we did not try MLM fine-tuning using the dialect version of NDC to produce an expert for Layer Swapping; on inspection of the corpus, the Bokmål version seemed closer to the target language, and given the unpromising results using the other MLM experts we did not explore this further.

\section*{Acknowledgements}
We thank the anonymous reviewers for their suggestions.
This research is supported by European Research Council (ERC) Consolidator Grant DIALECT 101043235.

\bibliography{anthology,custom}

\begin{thebibliography}{50}
\providecommand{\natexlab}[1]{#1}

\bibitem[{Abboud and Oz(2024)}]{abboud-oz-2024-towards}
Khadige Abboud and Gokmen Oz. 2024.
\newblock \href {https://aclanthology.org/2024.lrec-main.1433} {Towards equitable natural language understanding systems for dialectal cohorts: Debiasing training data}.
\newblock In \emph{Proceedings of the 2024 Joint International Conference on Computational Linguistics, Language Resources and Evaluation (LREC-COLING 2024)}, pages 16487--16499, Torino, Italia. ELRA and ICCL.

\bibitem[{Aepli et~al.(2023)Aepli, {\c{C}}{\"o}ltekin, Van Der~Goot, Jauhiainen, Kazzaz, Ljube{\v{s}}i{\'c}, North, Plank, Scherrer, and Zampieri}]{aepli-etal-2023-findings}
No{\"e}mi Aepli, {\c{C}}a{\u{g}}r{\i} {\c{C}}{\"o}ltekin, Rob Van Der~Goot, Tommi Jauhiainen, Mourhaf Kazzaz, Nikola Ljube{\v{s}}i{\'c}, Kai North, Barbara Plank, Yves Scherrer, and Marcos Zampieri. 2023.
\newblock \href {https://doi.org/10.18653/v1/2023.vardial-1.25} {Findings of the {V}ar{D}ial evaluation campaign 2023}.
\newblock In \emph{Tenth Workshop on NLP for Similar Languages, Varieties and Dialects (VarDial 2023)}, pages 251--261, Dubrovnik, Croatia. Association for Computational Linguistics.

\bibitem[{Aepli and Sennrich(2022)}]{aepli-sennrich-2022-improving}
No{\"e}mi Aepli and Rico Sennrich. 2022.
\newblock \href {https://doi.org/10.18653/v1/2022.findings-acl.321} {Improving zero-shot cross-lingual transfer between closely related languages by injecting character-level noise}.
\newblock In \emph{Findings of the Association for Computational Linguistics: ACL 2022}, pages 4074--4083, Dublin, Ireland. Association for Computational Linguistics.

\bibitem[{Artemova et~al.(2024)Artemova, Blaschke, and Plank}]{artemova-etal-2024-exploring}
Ekaterina Artemova, Verena Blaschke, and Barbara Plank. 2024.
\newblock \href {https://aclanthology.org/2024.eacl-long.28} {Exploring the robustness of task-oriented dialogue systems for colloquial {G}erman varieties}.
\newblock In \emph{Proceedings of the 18th Conference of the European Chapter of the Association for Computational Linguistics (Volume 1: Long Papers)}, pages 445--468, St. Julian{'}s, Malta. Association for Computational Linguistics.

\bibitem[{Bandarkar et~al.(2024)Bandarkar, Muller, Yuvraj, Hou, Singhal, Lv, and Liu}]{bandarkar2024layerswappingzeroshotcrosslingual}
Lucas Bandarkar, Benjamin Muller, Pritish Yuvraj, Rui Hou, Nayan Singhal, Hongjiang Lv, and Bing Liu. 2024.
\newblock \href {https://arxiv.org/abs/2410.01335} {Layer swapping for zero-shot cross-lingual transfer in large language models}.
\newblock \emph{Preprint}, arXiv:2410.01335.

\bibitem[{Blaschke et~al.(2024)Blaschke, Kova{\v{c}}i{\'c}, Peng, Sch{\"u}tze, and Plank}]{blaschke-etal-2024-maibaam}
Verena Blaschke, Barbara Kova{\v{c}}i{\'c}, Siyao Peng, Hinrich Sch{\"u}tze, and Barbara Plank. 2024.
\newblock \href {https://aclanthology.org/2024.lrec-main.953} {{M}ai{B}aam: A multi-dialectal {B}avarian {U}niversal {D}ependency treebank}.
\newblock In \emph{Proceedings of the 2024 Joint International Conference on Computational Linguistics, Language Resources and Evaluation (LREC-COLING 2024)}, pages 10921--10938, Torino, Italia. ELRA and ICCL.

\bibitem[{Blaschke et~al.(2023)Blaschke, Sch{\"u}tze, and Plank}]{blaschke-etal-2023-manipulating}
Verena Blaschke, Hinrich Sch{\"u}tze, and Barbara Plank. 2023.
\newblock \href {https://doi.org/10.18653/v1/2023.vardial-1.5} {Does manipulating tokenization aid cross-lingual transfer? a study on {POS} tagging for non-standardized languages}.
\newblock In \emph{Tenth Workshop on NLP for Similar Languages, Varieties and Dialects (VarDial 2023)}, pages 40--54, Dubrovnik, Croatia. Association for Computational Linguistics.

\bibitem[{Brahma et~al.(2023)Brahma, Maurya, and Desarkar}]{brahma-etal-2023-selectnoise}
Maharaj Brahma, Kaushal Maurya, and Maunendra Desarkar. 2023.
\newblock \href {https://doi.org/10.18653/v1/2023.findings-emnlp.109} {{S}elect{N}oise: Unsupervised noise injection to enable zero-shot machine translation for extremely low-resource languages}.
\newblock In \emph{Findings of the Association for Computational Linguistics: EMNLP 2023}, pages 1615--1629, Singapore. Association for Computational Linguistics.

\bibitem[{Chen et~al.(2019)Chen, Zhuo, and Wang}]{chen2019bertjointintentclassification}
Qian Chen, Zhu Zhuo, and Wen Wang. 2019.
\newblock \href {https://arxiv.org/abs/1902.10909} {{BERT} for joint intent classification and slot filling}.
\newblock \emph{Preprint}, arXiv:1902.10909.

\bibitem[{Conneau et~al.(2020)Conneau, Khandelwal, Goyal, Chaudhary, Wenzek, Guzm{\'a}n, Grave, Ott, Zettlemoyer, and Stoyanov}]{conneau-etal-2020-unsupervised}
Alexis Conneau, Kartikay Khandelwal, Naman Goyal, Vishrav Chaudhary, Guillaume Wenzek, Francisco Guzm{\'a}n, Edouard Grave, Myle Ott, Luke Zettlemoyer, and Veselin Stoyanov. 2020.
\newblock \href {https://doi.org/10.18653/v1/2020.acl-main.747} {Unsupervised cross-lingual representation learning at scale}.
\newblock In \emph{Proceedings of the 58th Annual Meeting of the Association for Computational Linguistics}, pages 8440--8451, Online. Association for Computational Linguistics.

\bibitem[{Coucke et~al.(2018)Coucke, Saade, Ball, Bluche, Caulier, Leroy, Doumouro, Gisselbrecht, Caltagirone, Lavril, Primet, and Dureau}]{coucke2018snips}
Alice Coucke, Alaa Saade, Adrien Ball, Théodore Bluche, Alexandre Caulier, David Leroy, Clément Doumouro, Thibault Gisselbrecht, Francesco Caltagirone, Thibaut Lavril, Maël Primet, and Joseph Dureau. 2018.
\newblock \href {https://arxiv.org/abs/1805.10190} {Snips voice platform: An embedded spoken language understanding system for private-by-design voice interfaces}.
\newblock \emph{Preprint}, arXiv:1805.10190.

\bibitem[{de~Marneffe et~al.(2021)de~Marneffe, Manning, Nivre, and Zeman}]{de-marneffe-etal-2021-universal}
Marie-Catherine de~Marneffe, Christopher~D. Manning, Joakim Nivre, and Daniel Zeman. 2021.
\newblock \href {https://doi.org/10.1162/coli_a_00402} {{U}niversal {D}ependencies}.
\newblock \emph{Computational Linguistics}, 47(2):255--308.

\bibitem[{Elkordi et~al.(2024)Elkordi, Sakr, Torki, and El-Makky}]{elkordi-etal-2024-alexunlp24}
Hossam Elkordi, Ahmed Sakr, Marwan Torki, and Nagwa El-Makky. 2024.
\newblock \href {https://doi.org/10.18653/v1/2024.arabicnlp-1.37} {{A}lexu{NLP}24 at {A}ra{F}in{NLP}2024: Multi-dialect {A}rabic intent detection with contrastive learning in banking domain}.
\newblock In \emph{Proceedings of The Second Arabic Natural Language Processing Conference}, pages 415--421, Bangkok, Thailand. Association for Computational Linguistics.

\bibitem[{Fares and Touileb(2024)}]{fares-touileb-2024-babelbot}
Murhaf Fares and Samia Touileb. 2024.
\newblock \href {https://doi.org/10.18653/v1/2024.arabicnlp-1.40} {{B}abel{B}ot at {A}ra{F}in{NLP}2024: Fine-tuning t5 for multi-dialect intent detection with synthetic data and model ensembling}.
\newblock In \emph{Proceedings of The Second Arabic Natural Language Processing Conference}, pages 433--440, Bangkok, Thailand. Association for Computational Linguistics.

\bibitem[{FitzGerald et~al.(2023)FitzGerald, Hench, Peris, Mackie, Rottmann, Sanchez, Nash, Urbach, Kakarala, Singh, Ranganath, Crist, Britan, Leeuwis, Tur, and Natarajan}]{fitzgerald-etal-2023-massive}
Jack FitzGerald, Christopher Hench, Charith Peris, Scott Mackie, Kay Rottmann, Ana Sanchez, Aaron Nash, Liam Urbach, Vishesh Kakarala, Richa Singh, Swetha Ranganath, Laurie Crist, Misha Britan, Wouter Leeuwis, Gokhan Tur, and Prem Natarajan. 2023.
\newblock \href {https://doi.org/10.18653/v1/2023.acl-long.235} {{MASSIVE}: A 1{M}-example multilingual natural language understanding dataset with 51 typologically-diverse languages}.
\newblock In \emph{Proceedings of the 61st Annual Meeting of the Association for Computational Linguistics (Volume 1: Long Papers)}, pages 4277--4302, Toronto, Canada. Association for Computational Linguistics.

\bibitem[{Grattafiori et~al.(2024)Grattafiori, Dubey, Jauhri, Pandey, Kadian, Al-Dahle, Letman, Mathur, Schelten, Vaughan, Yang, Fan, Goyal, Hartshorn, Yang, Mitra, Sravankumar, Korenev, Hinsvark, Rao, Zhang, Rodriguez, Gregerson, Spataru, Roziere, Biron, Tang, Chern, Caucheteux, Nayak, Bi, Marra, McConnell, Keller, Touret, Wu, Wong, Ferrer, Nikolaidis, Allonsius, Song, Pintz, Livshits, Wyatt, Esiobu, Choudhary, Mahajan, Garcia-Olano, Perino, Hupkes, Lakomkin, AlBadawy, Lobanova, Dinan, Smith, Radenovic, Guzmán, Zhang, Synnaeve, Lee, Anderson, Thattai, Nail, Mialon, Pang, Cucurell, Nguyen, Korevaar, Xu, Touvron, Zarov, Ibarra, Kloumann, Misra, Evtimov, Zhang, Copet, Lee, Geffert, Vranes, Park, Mahadeokar, Shah, van~der Linde, Billock, Hong, Lee, Fu, Chi, Huang, Liu, Wang, Yu, Bitton, Spisak, Park, Rocca, Johnstun, Saxe, Jia, Alwala, Prasad, Upasani, Plawiak, Li, Heafield, Stone, El-Arini, Iyer, Malik, Chiu, Bhalla, Lakhotia, Rantala-Yeary, van~der Maaten, Chen, Tan, Jenkins, Martin, Madaan, Malo, Blecher,
  Landzaat, de~Oliveira, Muzzi, Pasupuleti, Singh, Paluri, Kardas, Tsimpoukelli, Oldham, Rita, Pavlova, Kambadur, Lewis, Si, Singh, Hassan, Goyal, Torabi, Bashlykov, Bogoychev, Chatterji, Zhang, Duchenne, Çelebi, Alrassy, Zhang, Li, Vasic, Weng, Bhargava, Dubal, Krishnan, Koura, Xu, He, Dong, Srinivasan, Ganapathy, Calderer, Cabral, Stojnic, Raileanu, Maheswari, Girdhar, Patel, Sauvestre, Polidoro, Sumbaly, Taylor, Silva, Hou, Wang, Hosseini, Chennabasappa, Singh, Bell, Kim, Edunov, Nie, Narang, Raparthy, Shen, Wan, Bhosale, Zhang, Vandenhende, Batra, Whitman, Sootla, Collot, Gururangan, Borodinsky, Herman, Fowler, Sheasha, Georgiou, Scialom, Speckbacher, Mihaylov, Xiao, Karn, Goswami, Gupta, Ramanathan, Kerkez, Gonguet, Do, Vogeti, Albiero, Petrovic, Chu, Xiong, Fu, Meers, Martinet, Wang, Wang, Tan, Xia, Xie, Jia, Wang, Goldschlag, Gaur, Babaei, Wen, Song, Zhang, Li, Mao, Coudert, Yan, Chen, Papakipos, Singh, Srivastava, Jain, Kelsey, Shajnfeld, Gangidi, Victoria, Goldstand, Menon, Sharma, Boesenberg,
  Baevski, Feinstein, Kallet, Sangani, Teo, Yunus, Lupu, Alvarado, Caples, Gu, Ho, Poulton, Ryan, Ramchandani, Dong, Franco, Goyal, Saraf, Chowdhury, Gabriel, Bharambe, Eisenman, Yazdan, James, Maurer, Leonhardi, Huang, Loyd, Paola, Paranjape, Liu, Wu, Ni, Hancock, Wasti, Spence, Stojkovic, Gamido, Montalvo, Parker, Burton, Mejia, Liu, Wang, Kim, Zhou, Hu, Chu, Cai, Tindal, Feichtenhofer, Gao, Civin, Beaty, Kreymer, Li, Adkins, Xu, Testuggine, David, Parikh, Liskovich, Foss, Wang, Le, Holland, Dowling, Jamil, Montgomery, Presani, Hahn, Wood, Le, Brinkman, Arcaute, Dunbar, Smothers, Sun, Kreuk, Tian, Kokkinos, Ozgenel, Caggioni, Kanayet, Seide, Florez, Schwarz, Badeer, Swee, Halpern, Herman, Sizov, Guangyi, Zhang, Lakshminarayanan, Inan, Shojanazeri, Zou, Wang, Zha, Habeeb, Rudolph, Suk, Aspegren, Goldman, Zhan, Damlaj, Molybog, Tufanov, Leontiadis, Veliche, Gat, Weissman, Geboski, Kohli, Lam, Asher, Gaya, Marcus, Tang, Chan, Zhen, Reizenstein, Teboul, Zhong, Jin, Yang, Cummings, Carvill, Shepard, McPhie,
  Torres, Ginsburg, Wang, Wu, U, Saxena, Khandelwal, Zand, Matosich, Veeraraghavan, Michelena, Li, Jagadeesh, Huang, Chawla, Huang, Chen, Garg, A, Silva, Bell, Zhang, Guo, Yu, Moshkovich, Wehrstedt, Khabsa, Avalani, Bhatt, Mankus, Hasson, Lennie, Reso, Groshev, Naumov, Lathi, Keneally, Liu, Seltzer, Valko, Restrepo, Patel, Vyatskov, Samvelyan, Clark, Macey, Wang, Hermoso, Metanat, Rastegari, Bansal, Santhanam, Parks, White, Bawa, Singhal, Egebo, Usunier, Mehta, Laptev, Dong, Cheng, Chernoguz, Hart, Salpekar, Kalinli, Kent, Parekh, Saab, Balaji, Rittner, Bontrager, Roux, Dollar, Zvyagina, Ratanchandani, Yuvraj, Liang, Alao, Rodriguez, Ayub, Murthy, Nayani, Mitra, Parthasarathy, Li, Hogan, Battey, Wang, Howes, Rinott, Mehta, Siby, Bondu, Datta, Chugh, Hunt, Dhillon, Sidorov, Pan, Mahajan, Verma, Yamamoto, Ramaswamy, Lindsay, Lindsay, Feng, Lin, Zha, Patil, Shankar, Zhang, Zhang, Wang, Agarwal, Sajuyigbe, Chintala, Max, Chen, Kehoe, Satterfield, Govindaprasad, Gupta, Deng, Cho, Virk, Subramanian, Choudhury,
  Goldman, Remez, Glaser, Best, Koehler, Robinson, Li, Zhang, Matthews, Chou, Shaked, Vontimitta, Ajayi, Montanez, Mohan, Kumar, Mangla, Ionescu, Poenaru, Mihailescu, Ivanov, Li, Wang, Jiang, Bouaziz, Constable, Tang, Wu, Wang, Wu, Gao, Kleinman, Chen, Hu, Jia, Qi, Li, Zhang, Zhang, Adi, Nam, Yu, Wang, Zhao, Hao, Qian, Li, He, Rait, DeVito, Rosnbrick, Wen, Yang, Zhao, and Ma}]{grattafiori2024llama3herdmodels}
Aaron Grattafiori, Abhimanyu Dubey, Abhinav Jauhri, Abhinav Pandey, Abhishek Kadian, Ahmad Al-Dahle, Aiesha Letman, Akhil Mathur, Alan Schelten, Alex Vaughan, Amy Yang, Angela Fan, Anirudh Goyal, Anthony Hartshorn, Aobo Yang, Archi Mitra, Archie Sravankumar, Artem Korenev, Arthur Hinsvark, Arun Rao, Aston Zhang, Aurelien Rodriguez, Austen Gregerson, Ava Spataru, Baptiste Roziere, Bethany Biron, Binh Tang, Bobbie Chern, Charlotte Caucheteux, Chaya Nayak, Chloe Bi, Chris Marra, Chris McConnell, Christian Keller, Christophe Touret, Chunyang Wu, Corinne Wong, Cristian~Canton Ferrer, Cyrus Nikolaidis, Damien Allonsius, Daniel Song, Danielle Pintz, Danny Livshits, Danny Wyatt, David Esiobu, Dhruv Choudhary, Dhruv Mahajan, Diego Garcia-Olano, Diego Perino, Dieuwke Hupkes, Egor Lakomkin, Ehab AlBadawy, Elina Lobanova, Emily Dinan, Eric~Michael Smith, Filip Radenovic, Francisco Guzmán, Frank Zhang, Gabriel Synnaeve, Gabrielle Lee, Georgia~Lewis Anderson, Govind Thattai, Graeme Nail, Gregoire Mialon, Guan Pang,
  Guillem Cucurell, Hailey Nguyen, Hannah Korevaar, Hu~Xu, Hugo Touvron, Iliyan Zarov, Imanol~Arrieta Ibarra, Isabel Kloumann, Ishan Misra, Ivan Evtimov, Jack Zhang, Jade Copet, Jaewon Lee, Jan Geffert, Jana Vranes, Jason Park, Jay Mahadeokar, Jeet Shah, Jelmer van~der Linde, Jennifer Billock, Jenny Hong, Jenya Lee, Jeremy Fu, Jianfeng Chi, Jianyu Huang, Jiawen Liu, Jie Wang, Jiecao Yu, Joanna Bitton, Joe Spisak, Jongsoo Park, Joseph Rocca, Joshua Johnstun, Joshua Saxe, Junteng Jia, Kalyan~Vasuden Alwala, Karthik Prasad, Kartikeya Upasani, Kate Plawiak, Ke~Li, Kenneth Heafield, Kevin Stone, Khalid El-Arini, Krithika Iyer, Kshitiz Malik, Kuenley Chiu, Kunal Bhalla, Kushal Lakhotia, Lauren Rantala-Yeary, Laurens van~der Maaten, Lawrence Chen, Liang Tan, Liz Jenkins, Louis Martin, Lovish Madaan, Lubo Malo, Lukas Blecher, Lukas Landzaat, Luke de~Oliveira, Madeline Muzzi, Mahesh Pasupuleti, Mannat Singh, Manohar Paluri, Marcin Kardas, Maria Tsimpoukelli, Mathew Oldham, Mathieu Rita, Maya Pavlova, Melanie Kambadur,
  Mike Lewis, Min Si, Mitesh~Kumar Singh, Mona Hassan, Naman Goyal, Narjes Torabi, Nikolay Bashlykov, Nikolay Bogoychev, Niladri Chatterji, Ning Zhang, Olivier Duchenne, Onur Çelebi, Patrick Alrassy, Pengchuan Zhang, Pengwei Li, Petar Vasic, Peter Weng, Prajjwal Bhargava, Pratik Dubal, Praveen Krishnan, Punit~Singh Koura, Puxin Xu, Qing He, Qingxiao Dong, Ragavan Srinivasan, Raj Ganapathy, Ramon Calderer, Ricardo~Silveira Cabral, Robert Stojnic, Roberta Raileanu, Rohan Maheswari, Rohit Girdhar, Rohit Patel, Romain Sauvestre, Ronnie Polidoro, Roshan Sumbaly, Ross Taylor, Ruan Silva, Rui Hou, Rui Wang, Saghar Hosseini, Sahana Chennabasappa, Sanjay Singh, Sean Bell, Seohyun~Sonia Kim, Sergey Edunov, Shaoliang Nie, Sharan Narang, Sharath Raparthy, Sheng Shen, Shengye Wan, Shruti Bhosale, Shun Zhang, Simon Vandenhende, Soumya Batra, Spencer Whitman, Sten Sootla, Stephane Collot, Suchin Gururangan, Sydney Borodinsky, Tamar Herman, Tara Fowler, Tarek Sheasha, Thomas Georgiou, Thomas Scialom, Tobias Speckbacher,
  Todor Mihaylov, Tong Xiao, Ujjwal Karn, Vedanuj Goswami, Vibhor Gupta, Vignesh Ramanathan, Viktor Kerkez, Vincent Gonguet, Virginie Do, Vish Vogeti, Vítor Albiero, Vladan Petrovic, Weiwei Chu, Wenhan Xiong, Wenyin Fu, Whitney Meers, Xavier Martinet, Xiaodong Wang, Xiaofang Wang, Xiaoqing~Ellen Tan, Xide Xia, Xinfeng Xie, Xuchao Jia, Xuewei Wang, Yaelle Goldschlag, Yashesh Gaur, Yasmine Babaei, Yi~Wen, Yiwen Song, Yuchen Zhang, Yue Li, Yuning Mao, Zacharie~Delpierre Coudert, Zheng Yan, Zhengxing Chen, Zoe Papakipos, Aaditya Singh, Aayushi Srivastava, Abha Jain, Adam Kelsey, Adam Shajnfeld, Adithya Gangidi, Adolfo Victoria, Ahuva Goldstand, Ajay Menon, Ajay Sharma, Alex Boesenberg, Alexei Baevski, Allie Feinstein, Amanda Kallet, Amit Sangani, Amos Teo, Anam Yunus, Andrei Lupu, Andres Alvarado, Andrew Caples, Andrew Gu, Andrew Ho, Andrew Poulton, Andrew Ryan, Ankit Ramchandani, Annie Dong, Annie Franco, Anuj Goyal, Aparajita Saraf, Arkabandhu Chowdhury, Ashley Gabriel, Ashwin Bharambe, Assaf Eisenman, Azadeh
  Yazdan, Beau James, Ben Maurer, Benjamin Leonhardi, Bernie Huang, Beth Loyd, Beto~De Paola, Bhargavi Paranjape, Bing Liu, Bo~Wu, Boyu Ni, Braden Hancock, Bram Wasti, Brandon Spence, Brani Stojkovic, Brian Gamido, Britt Montalvo, Carl Parker, Carly Burton, Catalina Mejia, Ce~Liu, Changhan Wang, Changkyu Kim, Chao Zhou, Chester Hu, Ching-Hsiang Chu, Chris Cai, Chris Tindal, Christoph Feichtenhofer, Cynthia Gao, Damon Civin, Dana Beaty, Daniel Kreymer, Daniel Li, David Adkins, David Xu, Davide Testuggine, Delia David, Devi Parikh, Diana Liskovich, Didem Foss, Dingkang Wang, Duc Le, Dustin Holland, Edward Dowling, Eissa Jamil, Elaine Montgomery, Eleonora Presani, Emily Hahn, Emily Wood, Eric-Tuan Le, Erik Brinkman, Esteban Arcaute, Evan Dunbar, Evan Smothers, Fei Sun, Felix Kreuk, Feng Tian, Filippos Kokkinos, Firat Ozgenel, Francesco Caggioni, Frank Kanayet, Frank Seide, Gabriela~Medina Florez, Gabriella Schwarz, Gada Badeer, Georgia Swee, Gil Halpern, Grant Herman, Grigory Sizov, Guangyi, Zhang, Guna
  Lakshminarayanan, Hakan Inan, Hamid Shojanazeri, Han Zou, Hannah Wang, Hanwen Zha, Haroun Habeeb, Harrison Rudolph, Helen Suk, Henry Aspegren, Hunter Goldman, Hongyuan Zhan, Ibrahim Damlaj, Igor Molybog, Igor Tufanov, Ilias Leontiadis, Irina-Elena Veliche, Itai Gat, Jake Weissman, James Geboski, James Kohli, Janice Lam, Japhet Asher, Jean-Baptiste Gaya, Jeff Marcus, Jeff Tang, Jennifer Chan, Jenny Zhen, Jeremy Reizenstein, Jeremy Teboul, Jessica Zhong, Jian Jin, Jingyi Yang, Joe Cummings, Jon Carvill, Jon Shepard, Jonathan McPhie, Jonathan Torres, Josh Ginsburg, Junjie Wang, Kai Wu, Kam~Hou U, Karan Saxena, Kartikay Khandelwal, Katayoun Zand, Kathy Matosich, Kaushik Veeraraghavan, Kelly Michelena, Keqian Li, Kiran Jagadeesh, Kun Huang, Kunal Chawla, Kyle Huang, Lailin Chen, Lakshya Garg, Lavender A, Leandro Silva, Lee Bell, Lei Zhang, Liangpeng Guo, Licheng Yu, Liron Moshkovich, Luca Wehrstedt, Madian Khabsa, Manav Avalani, Manish Bhatt, Martynas Mankus, Matan Hasson, Matthew Lennie, Matthias Reso, Maxim
  Groshev, Maxim Naumov, Maya Lathi, Meghan Keneally, Miao Liu, Michael~L. Seltzer, Michal Valko, Michelle Restrepo, Mihir Patel, Mik Vyatskov, Mikayel Samvelyan, Mike Clark, Mike Macey, Mike Wang, Miquel~Jubert Hermoso, Mo~Metanat, Mohammad Rastegari, Munish Bansal, Nandhini Santhanam, Natascha Parks, Natasha White, Navyata Bawa, Nayan Singhal, Nick Egebo, Nicolas Usunier, Nikhil Mehta, Nikolay~Pavlovich Laptev, Ning Dong, Norman Cheng, Oleg Chernoguz, Olivia Hart, Omkar Salpekar, Ozlem Kalinli, Parkin Kent, Parth Parekh, Paul Saab, Pavan Balaji, Pedro Rittner, Philip Bontrager, Pierre Roux, Piotr Dollar, Polina Zvyagina, Prashant Ratanchandani, Pritish Yuvraj, Qian Liang, Rachad Alao, Rachel Rodriguez, Rafi Ayub, Raghotham Murthy, Raghu Nayani, Rahul Mitra, Rangaprabhu Parthasarathy, Raymond Li, Rebekkah Hogan, Robin Battey, Rocky Wang, Russ Howes, Ruty Rinott, Sachin Mehta, Sachin Siby, Sai~Jayesh Bondu, Samyak Datta, Sara Chugh, Sara Hunt, Sargun Dhillon, Sasha Sidorov, Satadru Pan, Saurabh Mahajan,
  Saurabh Verma, Seiji Yamamoto, Sharadh Ramaswamy, Shaun Lindsay, Shaun Lindsay, Sheng Feng, Shenghao Lin, Shengxin~Cindy Zha, Shishir Patil, Shiva Shankar, Shuqiang Zhang, Shuqiang Zhang, Sinong Wang, Sneha Agarwal, Soji Sajuyigbe, Soumith Chintala, Stephanie Max, Stephen Chen, Steve Kehoe, Steve Satterfield, Sudarshan Govindaprasad, Sumit Gupta, Summer Deng, Sungmin Cho, Sunny Virk, Suraj Subramanian, Sy~Choudhury, Sydney Goldman, Tal Remez, Tamar Glaser, Tamara Best, Thilo Koehler, Thomas Robinson, Tianhe Li, Tianjun Zhang, Tim Matthews, Timothy Chou, Tzook Shaked, Varun Vontimitta, Victoria Ajayi, Victoria Montanez, Vijai Mohan, Vinay~Satish Kumar, Vishal Mangla, Vlad Ionescu, Vlad Poenaru, Vlad~Tiberiu Mihailescu, Vladimir Ivanov, Wei Li, Wenchen Wang, Wenwen Jiang, Wes Bouaziz, Will Constable, Xiaocheng Tang, Xiaojian Wu, Xiaolan Wang, Xilun Wu, Xinbo Gao, Yaniv Kleinman, Yanjun Chen, Ye~Hu, Ye~Jia, Ye~Qi, Yenda Li, Yilin Zhang, Ying Zhang, Yossi Adi, Youngjin Nam, Yu, Wang, Yu~Zhao, Yuchen Hao, Yundi
  Qian, Yunlu Li, Yuzi He, Zach Rait, Zachary DeVito, Zef Rosnbrick, Zhaoduo Wen, Zhenyu Yang, Zhiwei Zhao, and Zhiyu Ma. 2024.
\newblock \href {https://arxiv.org/abs/2407.21783} {The llama 3 herd of models}.
\newblock \emph{Preprint}, arXiv:2407.21783.

\bibitem[{Hagen et~al.(2018)Hagen, Håberg, Olsen, and Søfteland}]{lia-transkripsjon}
Kristin Hagen, Live Håberg, Eirik Olsen, and Åshild Søfteland. 2018.
\newblock \href {https://tekstlab.uio.no/LIA/pdf/transkripsjonsrettleiing_lia.pdf} {Transkripsjonsrettleiing for {LIA}}.

\bibitem[{He et~al.(2023)He, Gao, and Chen}]{he2023debertav3}
Pengcheng He, Jianfeng Gao, and Weizhu Chen. 2023.
\newblock \href {https://openreview.net/pdf?id=sE7-XhLxHA} {{DeBERTaV3}: Improving {DeBERTa} using {ELECTRA}-style pre-training with gradient-disentangled embedding sharing}.
\newblock In \emph{Proceedings of the Eleventh International Conference on Learning Representations (ICLR)}.

\bibitem[{He et~al.(2021)He, Liu, Gao, and Chen}]{he2021deberta}
Pengcheng He, Xiaodong Liu, Jianfeng Gao, and Weizhu Chen. 2021.
\newblock \href {https://openreview.net/forum?id=XPZIaotutsD} {{DeBERTa}: Decoding-enhanced {BERT} with disentangled attention}.
\newblock In \emph{International Conference on Learning Representations}.

\bibitem[{Hedderich et~al.(2021)Hedderich, Lange, Adel, Str{\"o}tgen, and Klakow}]{hedderich-etal-2021-survey}
Michael~A. Hedderich, Lukas Lange, Heike Adel, Jannik Str{\"o}tgen, and Dietrich Klakow. 2021.
\newblock \href {https://doi.org/10.18653/v1/2021.naacl-main.201} {A survey on recent approaches for natural language processing in low-resource scenarios}.
\newblock In \emph{Proceedings of the 2021 Conference of the North American Chapter of the Association for Computational Linguistics: Human Language Technologies}, pages 2545--2568, Online. Association for Computational Linguistics.

\bibitem[{Johannessen et~al.(2009)Johannessen, Priestley, Hagen, {\AA}farli, and Vangsnes}]{johannessen-etal-2009-nordic}
Janne~Bondi Johannessen, Joel~James Priestley, Kristin Hagen, Tor~Anders {\AA}farli, and {\O}ystein~Alexander Vangsnes. 2009.
\newblock \href {https://aclanthology.org/W09-4612} {The {N}ordic dialect corpus{--}an advanced research tool}.
\newblock In \emph{Proceedings of the 17th Nordic Conference of Computational Linguistics ({NODALIDA} 2009)}, pages 73--80, Odense, Denmark. Northern European Association for Language Technology (NEALT).

\bibitem[{J{\o}rgensen et~al.(2020)J{\o}rgensen, Aasmoe, Ruud~Husev{\aa}g, {\O}vrelid, and Velldal}]{jorgensen-etal-2020-norne}
Fredrik J{\o}rgensen, Tobias Aasmoe, Anne-Stine Ruud~Husev{\aa}g, Lilja {\O}vrelid, and Erik Velldal. 2020.
\newblock \href {https://aclanthology.org/2020.lrec-1.559} {{N}or{NE}: Annotating named entities for {N}orwegian}.
\newblock In \emph{Proceedings of the Twelfth Language Resources and Evaluation Conference}, pages 4547--4556, Marseille, France. European Language Resources Association.

\bibitem[{Krückl et~al.(2025)Krückl, Blaschke, and Plank}]{krueckl2025improving}
Xaver~Maria Krückl, Verena Blaschke, and Barbara Plank. 2025.
\newblock Improving dialectal slot and intent detection with auxiliary tasks: {A} multi-dialectal {Bavarian} case study.
\newblock In \emph{Proceedings of the Twelfth Workshop on NLP for Similar Languages, Varieties, and Dialects (VarDial 2025)}, Abu Dhabi, UAE. International Committee on Computational Linguistics.

\bibitem[{Kwon et~al.(2023)Kwon, Bhatia, Nagoudi, Alcoba~Inciarte, and Abdul-mageed}]{kwon-etal-2023-sidlr}
Sang~Yun Kwon, Gagan Bhatia, Elmoatez~Billah Nagoudi, Alcides Alcoba~Inciarte, and Muhammad Abdul-mageed. 2023.
\newblock \href {https://doi.org/10.18653/v1/2023.vardial-1.24} {{SIDLR}: Slot and intent detection models for low-resource language varieties}.
\newblock In \emph{Tenth Workshop on NLP for Similar Languages, Varieties and Dialects (VarDial 2023)}, pages 241--250, Dubrovnik, Croatia. Association for Computational Linguistics.

\bibitem[{Lafferty et~al.(2001)Lafferty, McCallum, Pereira et~al.}]{lafferty2001conditional}
John Lafferty, Andrew McCallum, Fernando Pereira, et~al. 2001.
\newblock Conditional random fields: Probabilistic models for segmenting and labeling sequence data.
\newblock In \emph{Proceedings of the Eighteenth International Conference on Machine Learning}, pages 282--289, San Francisco, CA, USA. Morgan Kaufmann Publishers Inc.

\bibitem[{Li et~al.(2021)Li, Arora, Chen, Gupta, Gupta, and Mehdad}]{li-etal-2021-mtop}
Haoran Li, Abhinav Arora, Shuohui Chen, Anchit Gupta, Sonal Gupta, and Yashar Mehdad. 2021.
\newblock \href {https://doi.org/10.18653/v1/2021.eacl-main.257} {{MTOP}: A comprehensive multilingual task-oriented semantic parsing benchmark}.
\newblock In \emph{Proceedings of the 16th Conference of the European Chapter of the Association for Computational Linguistics: Main Volume}, pages 2950--2962, Online. Association for Computational Linguistics.

\bibitem[{M{\ae}hlum and Scherrer(2024)}]{maehlum-scherrer-2024-nomusic}
Petter M{\ae}hlum and Yves Scherrer. 2024.
\newblock \href {https://doi.org/10.18653/v1/2024.vardial-1.9} {{N}o{M}usic - the {N}orwegian multi-dialectal slot and intent detection corpus}.
\newblock In \emph{Proceedings of the Eleventh Workshop on NLP for Similar Languages, Varieties, and Dialects (VarDial 2024)}, pages 107--116, Mexico City, Mexico. Association for Computational Linguistics.

\bibitem[{Malaysha et~al.(2024)Malaysha, El-Haj, Ezzini, Khalilia, Jarrar, Almujaiwel, Berrada, and Bouamor}]{malaysha-etal-2024-arafinnlp}
Sanad Malaysha, Mo~El-Haj, Saad Ezzini, Mohammed Khalilia, Mustafa Jarrar, Sultan Almujaiwel, Ismail Berrada, and Houda Bouamor. 2024.
\newblock \href {https://doi.org/10.18653/v1/2024.arabicnlp-1.34} {{A}ra{F}in{NLP} 2024: The first {A}rabic financial {NLP} shared task}.
\newblock In \emph{Proceedings of The Second Arabic Natural Language Processing Conference}, pages 393--402, Bangkok, Thailand. Association for Computational Linguistics.

\bibitem[{Montariol et~al.(2022)Montariol, Riabi, and Seddah}]{montariol-etal-2022-multilingual}
Syrielle Montariol, Arij Riabi, and Djam{\'e} Seddah. 2022.
\newblock \href {https://aclanthology.org/2022.findings-aacl.33} {Multilingual auxiliary tasks training: Bridging the gap between languages for zero-shot transfer of hate speech detection models}.
\newblock In \emph{Findings of the Association for Computational Linguistics: AACL-IJCNLP 2022}, pages 347--363, Online only. Association for Computational Linguistics.

\bibitem[{Mu{\~n}oz-Ortiz et~al.(2025)Mu{\~n}oz-Ortiz, Blaschke, and Plank}]{munoz-ortiz-etal-2025-evaluating}
Alberto Mu{\~n}oz-Ortiz, Verena Blaschke, and Barbara Plank. 2025.
\newblock \href {https://arxiv.org/abs/2412.09084} {Evaluating pixel language models on non-standardized languages}.
\newblock In \emph{Proceedings of the 31st International Conference on Computational Linguistics (COLING 2025)}, Abu Dhabi, UAE. International Committee on Computational Linguistics.

\bibitem[{{\O}vrelid et~al.(2018){\O}vrelid, K{\aa}sen, Hagen, N{\o}klestad, Solberg, and Johannessen}]{ovrelid-etal-2018-lia}
Lilja {\O}vrelid, Andre K{\aa}sen, Kristin Hagen, Anders N{\o}klestad, Per~Erik Solberg, and Janne~Bondi Johannessen. 2018.
\newblock \href {https://aclanthology.org/L18-1710} {The {LIA} treebank of spoken {N}orwegian dialects}.
\newblock In \emph{Proceedings of the Eleventh International Conference on Language Resources and Evaluation ({LREC} 2018)}, Miyazaki, Japan. European Language Resources Association (ELRA).

\bibitem[{Padmakumar et~al.(2022)Padmakumar, Lausen, Ballesteros, Zha, He, and Karypis}]{padmakumar-etal-2022-exploring}
Vishakh Padmakumar, Leonard Lausen, Miguel Ballesteros, Sheng Zha, He~He, and George Karypis. 2022.
\newblock \href {https://doi.org/10.18653/v1/2022.naacl-main.183} {Exploring the role of task transferability in large-scale multi-task learning}.
\newblock In \emph{Proceedings of the 2022 Conference of the North American Chapter of the Association for Computational Linguistics: Human Language Technologies}, pages 2542--2550, Seattle, United States. Association for Computational Linguistics.

\bibitem[{Poth et~al.(2021)Poth, Pfeiffer, R{\"u}ckl{\'e}, and Gurevych}]{poth-etal-2021-pre}
Clifton Poth, Jonas Pfeiffer, Andreas R{\"u}ckl{\'e}, and Iryna Gurevych. 2021.
\newblock \href {https://doi.org/10.18653/v1/2021.emnlp-main.827} {{W}hat to pre-train on? {E}fficient intermediate task selection}.
\newblock In \emph{Proceedings of the 2021 Conference on Empirical Methods in Natural Language Processing}, pages 10585--10605, Online and Punta Cana, Dominican Republic. Association for Computational Linguistics.

\bibitem[{Pruksachatkun et~al.(2020)Pruksachatkun, Phang, Liu, Htut, Zhang, Pang, Vania, Kann, and Bowman}]{pruksachatkun-etal-2020-intermediate}
Yada Pruksachatkun, Jason Phang, Haokun Liu, Phu~Mon Htut, Xiaoyi Zhang, Richard~Yuanzhe Pang, Clara Vania, Katharina Kann, and Samuel~R. Bowman. 2020.
\newblock \href {https://doi.org/10.18653/v1/2020.acl-main.467} {Intermediate-task transfer learning with pretrained language models: When and why does it work?}
\newblock In \emph{Proceedings of the 58th Annual Meeting of the Association for Computational Linguistics}, pages 5231--5247, Online. Association for Computational Linguistics.

\bibitem[{Ramadan et~al.(2024)Ramadan, Amr, Torki, and El-Makky}]{ramadan-etal-2024-ma}
Asmaa Ramadan, Manar Amr, Marwan Torki, and Nagwa El-Makky. 2024.
\newblock \href {https://doi.org/10.18653/v1/2024.arabicnlp-1.41} {{MA} at {A}ra{F}in{NLP}2024: {BERT}-based ensemble for cross-dialectal {A}rabic intent detection}.
\newblock In \emph{Proceedings of The Second Arabic Natural Language Processing Conference}, pages 441--445, Bangkok, Thailand. Association for Computational Linguistics.

\bibitem[{Reimers and Gurevych(2019)}]{reimers-gurevych-2019-sentence}
Nils Reimers and Iryna Gurevych. 2019.
\newblock \href {https://doi.org/10.18653/v1/D19-1410} {Sentence-{BERT}: Sentence embeddings using {S}iamese {BERT}-networks}.
\newblock In \emph{Proceedings of the 2019 Conference on Empirical Methods in Natural Language Processing and the 9th International Joint Conference on Natural Language Processing (EMNLP-IJCNLP)}, pages 3982--3992, Hong Kong, China. Association for Computational Linguistics.

\bibitem[{Ruder(2017)}]{ruder2017overviewmultitasklearningdeep}
Sebastian Ruder. 2017.
\newblock \href {https://arxiv.org/abs/1706.05098} {An overview of multi-task learning in deep neural networks}.
\newblock \emph{Preprint}, arXiv:1706.05098.

\bibitem[{Samuel et~al.(2023)Samuel, Kutuzov, Touileb, Velldal, {\O}vrelid, R{\o}nningstad, Sigdel, and Palatkina}]{samuel-etal-2023-norbench}
David Samuel, Andrey Kutuzov, Samia Touileb, Erik Velldal, Lilja {\O}vrelid, Egil R{\o}nningstad, Elina Sigdel, and Anna Palatkina. 2023.
\newblock \href {https://aclanthology.org/2023.nodalida-1.61} {{N}or{B}ench {--} a benchmark for {N}orwegian language models}.
\newblock In \emph{Proceedings of the 24th Nordic Conference on Computational Linguistics (NoDaLiDa)}, pages 618--633, T{\'o}rshavn, Faroe Islands. University of Tartu Library.

\bibitem[{Scherrer et~al.(2025)Scherrer, van~der Goot, and M{\ae}hlum}]{scherrer-etal-2025-vardial}
Yves Scherrer, Rob van~der Goot, and Petter M{\ae}hlum. 2025.
\newblock {V}ar{D}ial evaluation campaign 2025: {N}orwegian slot and intent detection and dialect identification {(NorSID)}.
\newblock In \emph{Proceedings of the Twelfth Workshop on NLP for Similar Languages, Varieties, and Dialects (VarDial 2025)}, Abu Dhabi, UAE. International Committee on Computational Linguistics.

\bibitem[{Schr{\"o}der and Biemann(2020)}]{schroder-biemann-2020-estimating}
Fynn Schr{\"o}der and Chris Biemann. 2020.
\newblock \href {https://doi.org/10.18653/v1/2020.acl-main.268} {Estimating the influence of auxiliary tasks for multi-task learning of sequence tagging tasks}.
\newblock In \emph{Proceedings of the 58th Annual Meeting of the Association for Computational Linguistics}, pages 2971--2985, Online. Association for Computational Linguistics.

\bibitem[{Schuster et~al.(2019)Schuster, Gupta, Shah, and Lewis}]{schuster-etal-2019-cross-lingual}
Sebastian Schuster, Sonal Gupta, Rushin Shah, and Mike Lewis. 2019.
\newblock \href {https://doi.org/10.18653/v1/N19-1380} {Cross-lingual transfer learning for multilingual task oriented dialog}.
\newblock In \emph{Proceedings of the 2019 Conference of the North {A}merican Chapter of the Association for Computational Linguistics: Human Language Technologies, Volume 1 (Long and Short Papers)}, pages 3795--3805, Minneapolis, Minnesota. Association for Computational Linguistics.

\bibitem[{Sn{\ae}bjarnarson et~al.(2023)Sn{\ae}bjarnarson, Simonsen, Glava{\v{s}}, and Vuli{\'c}}]{snaebjarnarson-etal-2023-transfer}
V{\'e}steinn Sn{\ae}bjarnarson, Annika Simonsen, Goran Glava{\v{s}}, and Ivan Vuli{\'c}. 2023.
\newblock \href {https://aclanthology.org/2023.nodalida-1.74} {Transfer to a low-resource language via close relatives: The case study on {F}aroese}.
\newblock In \emph{Proceedings of the 24th Nordic Conference on Computational Linguistics (NoDaLiDa)}, pages 728--737, T{\'o}rshavn, Faroe Islands. University of Tartu Library.

\bibitem[{Solberg et~al.(2014)Solberg, Skj{\ae}rholt, {\O}vrelid, Hagen, and Johannessen}]{solberg-etal-2014-norwegian}
Per~Erik Solberg, Arne Skj{\ae}rholt, Lilja {\O}vrelid, Kristin Hagen, and Janne~Bondi Johannessen. 2014.
\newblock \href {http://www.lrec-conf.org/proceedings/lrec2014/pdf/303_Paper.pdf} {The {N}orwegian dependency treebank}.
\newblock In \emph{Proceedings of the Ninth International Conference on Language Resources and Evaluation ({LREC}'14)}, pages 789--795, Reykjavik, Iceland. European Language Resources Association (ELRA).

\bibitem[{Srivastava and Chiang(2023{\natexlab{a}})}]{srivastava-chiang-2023-bertwich}
Aarohi Srivastava and David Chiang. 2023{\natexlab{a}}.
\newblock \href {https://doi.org/10.18653/v1/2023.findings-emnlp.1037} {{BERT}wich: Extending {BERT}{'}s capabilities to model dialectal and noisy text}.
\newblock In \emph{Findings of the Association for Computational Linguistics: EMNLP 2023}, pages 15510--15521, Singapore. Association for Computational Linguistics.

\bibitem[{Srivastava and Chiang(2023{\natexlab{b}})}]{srivastava-chiang-2023-fine}
Aarohi Srivastava and David Chiang. 2023{\natexlab{b}}.
\newblock \href {https://doi.org/10.18653/v1/2023.vardial-1.16} {Fine-tuning {BERT} with character-level noise for zero-shot transfer to dialects and closely-related languages}.
\newblock In \emph{Tenth Workshop on NLP for Similar Languages, Varieties and Dialects (VarDial 2023)}, pages 152--162, Dubrovnik, Croatia. Association for Computational Linguistics.

\bibitem[{van~der Goot et~al.(2021{\natexlab{a}})van~der Goot, Sharaf, Imankulova, {\"U}st{\"u}n, Stepanovi{\'c}, Ramponi, Khairunnisa, Komachi, and Plank}]{van-der-goot-etal-2021-masked}
Rob van~der Goot, Ibrahim Sharaf, Aizhan Imankulova, Ahmet {\"U}st{\"u}n, Marija Stepanovi{\'c}, Alan Ramponi, Siti~Oryza Khairunnisa, Mamoru Komachi, and Barbara Plank. 2021{\natexlab{a}}.
\newblock \href {https://doi.org/10.18653/v1/2021.naacl-main.197} {From masked language modeling to translation: Non-{E}nglish auxiliary tasks improve zero-shot spoken language understanding}.
\newblock In \emph{Proceedings of the 2021 Conference of the North American Chapter of the Association for Computational Linguistics: Human Language Technologies}, pages 2479--2497, Online. Association for Computational Linguistics.

\bibitem[{van~der Goot et~al.(2021{\natexlab{b}})van~der Goot, {\"U}st{\"u}n, Ramponi, Sharaf, and Plank}]{van-der-goot-etal-2021-massive}
Rob van~der Goot, Ahmet {\"U}st{\"u}n, Alan Ramponi, Ibrahim Sharaf, and Barbara Plank. 2021{\natexlab{b}}.
\newblock \href {https://doi.org/10.18653/v1/2021.eacl-demos.22} {Massive choice, ample tasks ({M}a{C}h{A}mp): A toolkit for multi-task learning in {NLP}}.
\newblock In \emph{Proceedings of the 16th Conference of the European Chapter of the Association for Computational Linguistics: System Demonstrations}, pages 176--197, Online. Association for Computational Linguistics.

\bibitem[{Winkler et~al.(2024)Winkler, Juozapaityte, van~der Goot, and Plank}]{winkler-etal-2024-slot}
Miriam Winkler, Virginija Juozapaityte, Rob van~der Goot, and Barbara Plank. 2024.
\newblock \href {https://aclanthology.org/2024.lrec-main.1297} {Slot and intent detection resources for {B}avarian and {L}ithuanian: Assessing translations vs natural queries to digital assistants}.
\newblock In \emph{Proceedings of the 2024 Joint International Conference on Computational Linguistics, Language Resources and Evaluation (LREC-COLING 2024)}, pages 14898--14915, Torino, Italia. ELRA and ICCL.

\bibitem[{Xu et~al.(2020)Xu, Haider, and Mansour}]{xu-etal-2020-end}
Weijia Xu, Batool Haider, and Saab Mansour. 2020.
\newblock \href {https://doi.org/10.18653/v1/2020.emnlp-main.410} {End-to-end slot alignment and recognition for cross-lingual {NLU}}.
\newblock In \emph{Proceedings of the 2020 Conference on Empirical Methods in Natural Language Processing (EMNLP)}, pages 5052--5063, Online. Association for Computational Linguistics.

\bibitem[{Yao et~al.(2013)Yao, Zweig, Hwang, Shi, and Yu}]{yao13b_interspeech}
Kaisheng Yao, Geoffrey Zweig, Mei-Yuh Hwang, Yangyang Shi, and Dong Yu. 2013.
\newblock \href {https://doi.org/10.21437/Interspeech.2013-569} {Recurrent neural networks for language understanding}.
\newblock In \emph{Interspeech 2013}, pages 2524--2528.

\end{thebibliography}

\appendix

\section{Spelling Changes to the Dialectological Transcriptions}
\label{sec:appendix-spelling}

We make slight changes to the dialectological transcriptions used in LIA  based on LIA's transcription guidelines \cite{lia-transkripsjon}.
The idea is to turn the transcriptions into slightly more plausible spellings, but we want to stress that these rules are simplistic and not meant to produce text that fully emulates naturalistic dialect spellings.
\begin{itemize}
    \item We replace \ortho{L} (/\textrtailr/, \textit{tjukk l} `thick l') with \ortho{l}. While it can also correspond to \ortho{rd}, we found that it much more often corresponds to \ortho{l} in the data.
    \item We remove apostrophes (originally used to mark syllabic consonants).
    \item The dialectological transcriptions use double consonants to mark short vowels, which can lead to consonant clusters that are unlikely to occur in written Norwegian. In words where a double consonant is followed by at least one more consonant, we remove one of the doubled consonants (\textit{C\textsubscript{1}C\textsubscript{1}C\textsubscript{2}$\rightarrow$C\textsubscript{1}C\textsubscript{2}}). 
    If the sequence is \ortho{ssjt} or \ortho{ssjk}, we instead replace it with \ortho{rst} or \ortho{rsk}, respectively.
    If it otherwise starts with \ortho{ssj} or \ortho{kkj}, we do not remove the first \ortho{s} or \ortho{k}. 
\end{itemize}

\section{Detailed Results}
\label{sec:appendix-results}

\newcommand{\asteriskdev}{} %

\begin{table*}[t]
\centering
\setlength{\tabcolsep}{3pt}
\adjustbox{max width=\linewidth}{%
\begin{tabular}{@{}lllrrrlrrrl@{}}
\toprule
  &  & & \multicolumn{3}{c}{\textbf{Intents (acc., \%)}} &  & \multicolumn{3}{c}{\textbf{Slots (span \fonescore, \%)}}\\
\cmidrule(lr){4-6} \cmidrule(lr){8-10}
\textbf{Training data} & \textbf{PLM} & \textbf{Details} & \multicolumn{1}{c}{\textbf{Dev}} & \multicolumn{1}{c}{\textbf{Test}} & \multicolumn{2}{c}{\textbf{Subm.}}  & \multicolumn{1}{c}{\textbf{Dev}} & \multicolumn{1}{c}{\textbf{Test}} & \multicolumn{2}{c}{\textbf{Subm.}} \\
\midrule
English (train) & NorBERT & baseline & \cellcolor[HTML]{F0E194}96.9\,\textsubscript{0.4} & \cellcolor[HTML]{FFE9A6}95.1\,\textsubscript{0.2} &  &  & \cellcolor[HTML]{CAD687}79.9\,\textsubscript{0.1} & \cellcolor[HTML]{CCD688}79.7\,\textsubscript{0.4} \\
\horizontalspacer
 & ScandiBERT & baseline & \cellcolor[HTML]{FFE69C}96.4\,\textsubscript{0.5} & \cellcolor[HTML]{FFE9A8}94.8\,\textsubscript{0.8} &  &  & \cellcolor[HTML]{BED283}81.3\,\textsubscript{0.3} & \cellcolor[HTML]{C3D385}80.7\,\textsubscript{0.7} \\
  &  & dial\multitask{} & \cellcolor[HTML]{FFFFFC}84.3\,\textsubscript{2.7} & \cellcolor[HTML]{FFFFFF}83.8\,\textsubscript{3.2} &  &  & \cellcolor[HTML]{EFE194}75.8\,\textsubscript{1.8} & \cellcolor[HTML]{F0E194}75.8\,\textsubscript{1.2} \\
  &  & dial \sequential{} & \cellcolor[HTML]{FFE7A0}95.8\,\textsubscript{0.7} & \cellcolor[HTML]{FFEBAF}94.0\,\textsubscript{1.7} &  &  & \cellcolor[HTML]{CCD688}79.7\,\textsubscript{1.3} & \cellcolor[HTML]{D1D789}79.2\,\textsubscript{0.9} \\
  &  & POS\multitask{} & \cellcolor[HTML]{FFE79E}96.0\,\textsubscript{0.5} & \cellcolor[HTML]{FFE9A8}94.9\,\textsubscript{0.3} &  &  & \cellcolor[HTML]{BBD182}81.6\,\textsubscript{0.2} & \cellcolor[HTML]{BFD283}81.1\,\textsubscript{0.3} \\
  &  & POS \sequential{} & \cellcolor[HTML]{FFE69D}96.3\,\textsubscript{0.2} & \cellcolor[HTML]{FFEAA9}94.7\,\textsubscript{0.2} &  &  & \cellcolor[HTML]{B4CF80}82.3\,\textsubscript{0.8} & \cellcolor[HTML]{B6CF80}82.2\,\textsubscript{1.1} \\
  &  & dep\multitask{} & \cellcolor[HTML]{FFEAA9}94.7\,\textsubscript{1.2} & \cellcolor[HTML]{FFECB3}93.5\,\textsubscript{0.6} &  &  & \cellcolor[HTML]{B7D081}82.0\,\textsubscript{0.4} & \cellcolor[HTML]{BCD182}81.5\,\textsubscript{0.2} \\
  &  & dep \sequential{} & \cellcolor[HTML]{FFE599}96.7\,\textsubscript{1.0} & \cellcolor[HTML]{FFE9A8}94.9\,\textsubscript{1.2} &  &  & \cellcolor[HTML]{B3CE7F}82.5\,\textsubscript{0.9} & \cellcolor[HTML]{B9D081}81.8\,\textsubscript{0.7} & \cellcolor[HTML]{BAD081}82.6\\
 &  & NER\multitask{} & \cellcolor[HTML]{DBDB8D}97.2\,\textsubscript{0.9} & \cellcolor[HTML]{FFE8A4}95.3\,\textsubscript{1.0} &  &  & \cellcolor[HTML]{C1D384}80.9\,\textsubscript{0.9} & \cellcolor[HTML]{C4D485}80.6\,\textsubscript{1.0} \\
  &  & NER \sequential{} & \cellcolor[HTML]{F6E396}96.8\,\textsubscript{0.4} & \cellcolor[HTML]{FFE9A7}95.0\,\textsubscript{0.4} &  &  & \cellcolor[HTML]{BDD283}81.3\,\textsubscript{1.1} & \cellcolor[HTML]{BFD283}81.1\,\textsubscript{0.9} \\
 \horizontalspacer
 & mDeBERTa & baseline & \cellcolor[HTML]{FFE8A5}95.3\,\textsubscript{1.1} & \cellcolor[HTML]{FFEEBC}92.4\,\textsubscript{1.8} &  &  & \cellcolor[HTML]{E2DD8F}77.3\,\textsubscript{1.2} & \cellcolor[HTML]{E9DF92}76.5\,\textsubscript{1.2} \\
 \midrule
Norwegian (MT) & NorBERT & baseline & \cellcolor[HTML]{8DC372}{\color[HTML]{F3F3F3} 98.3\,\textsubscript{0.4}} & \cellcolor[HTML]{FFE69D}96.2\,\textsubscript{0.5} &  &  & \cellcolor[HTML]{FFFBED}55.7\,\textsubscript{0.5} & \cellcolor[HTML]{FFFDF5}53.9\,\textsubscript{0.3} \\
(train)  &  & noise (10\%) & \cellcolor[HTML]{83C06F}{\color[HTML]{F3F3F3} 98.5\,\textsubscript{0.4}} & \cellcolor[HTML]{FFE69B}96.4\,\textsubscript{0.3} &  &  & \cellcolor[HTML]{FFF9E6}57.3\,\textsubscript{0.2} & \cellcolor[HTML]{FFFCF0}55.1\,\textsubscript{0.8} \\
  &  & noise (20\%) & \cellcolor[HTML]{79BD6B}{\color[HTML]{F3F3F3} 98.6\,\textsubscript{0.2}} & \cellcolor[HTML]{DBDA8D}97.2\,\textsubscript{0.2} &  &  & \cellcolor[HTML]{FFFAEA}56.4\,\textsubscript{0.4} & \cellcolor[HTML]{FFFCF0}55.0\,\textsubscript{0.3} \\
  &  & noise (30\%) & \cellcolor[HTML]{5DB562}{\color[HTML]{F3F3F3} 99.0\,\textsubscript{0.1}} & \cellcolor[HTML]{CED788}97.4\,\textsubscript{0.5} &  &  & \cellcolor[HTML]{FFFAEA}56.4\,\textsubscript{1.3} & \cellcolor[HTML]{FFFDF5}54.1\,\textsubscript{1.1} \\
\horizontalspacer
 & ScandiBERT & baseline & \cellcolor[HTML]{BFD283}97.6\,\textsubscript{0.0} & \cellcolor[HTML]{FFE69C}96.3\,\textsubscript{0.1} &  &  & \cellcolor[HTML]{FFFBEE}55.5\,\textsubscript{0.4} & \cellcolor[HTML]{FFFCF2}54.6\,\textsubscript{0.4} \\
 &  & noise (10\%) & \cellcolor[HTML]{B2CE7F}97.8\,\textsubscript{0.1} & \cellcolor[HTML]{FFE69B}96.5\,\textsubscript{0.4} &  &  & \cellcolor[HTML]{FFF9E8}56.9\,\textsubscript{0.9} & \cellcolor[HTML]{FFFBEC}55.9\,\textsubscript{0.7} \\
  &  & noise (20\%) & \cellcolor[HTML]{84C16F}{\color[HTML]{F3F3F3} 98.4\,\textsubscript{0.3}} & \cellcolor[HTML]{C5D485}97.5\,\textsubscript{0.2} &  &  & \cellcolor[HTML]{FFFBEE}55.6\,\textsubscript{0.8} & \cellcolor[HTML]{FFFCF2}54.6\,\textsubscript{0.5} \\
  &  & noise (30\%) & \cellcolor[HTML]{A1C979}{\color[HTML]{F3F3F3} 98.0\,\textsubscript{0.4}} & \cellcolor[HTML]{E4DD90}97.1\,\textsubscript{0.5} &  &  & \cellcolor[HTML]{FFFAE9}56.5\,\textsubscript{0.6} & \cellcolor[HTML]{FFFCF1}54.8\,\textsubscript{0.5} \\
  &  & dial\multitask{} & \cellcolor[HTML]{FFF3D0}89.8\,\textsubscript{1.4} & \cellcolor[HTML]{FFF5D4}89.2\,\textsubscript{1.4} &  &  & \cellcolor[HTML]{FFFEFA}53.0\,\textsubscript{0.1} & \cellcolor[HTML]{FFFFFF}51.7\,\textsubscript{0.1} \\
  &  & dial \sequential{} & \cellcolor[HTML]{FFE79E}96.1\,\textsubscript{1.0} & \cellcolor[HTML]{FFE8A5}95.2\,\textsubscript{1.0} &  &  & \cellcolor[HTML]{FFFCF3}54.4\,\textsubscript{0.4} & \cellcolor[HTML]{FFFDF6}53.7\,\textsubscript{0.5} \\
  &  & POS\multitask{} & \cellcolor[HTML]{A8CB7B}97.9\,\textsubscript{0.3} & \cellcolor[HTML]{FAE498}96.8\,\textsubscript{0.4} &  &  & \cellcolor[HTML]{FFFDF5}54.0\,\textsubscript{0.5} & \cellcolor[HTML]{FFFDF6}53.7\,\textsubscript{0.6} \\
  &  & POS \sequential{} & \cellcolor[HTML]{B4CF7F}97.8\,\textsubscript{0.5} & \cellcolor[HTML]{FFE69A}96.7\,\textsubscript{0.4} &  &  & \cellcolor[HTML]{FFFBEE}55.5\,\textsubscript{0.4} & \cellcolor[HTML]{FFFCF3}54.4\,\textsubscript{0.8} \\
  &  & dep\multitask{} & \cellcolor[HTML]{A6CB7B}98.0\,\textsubscript{0.4} & \cellcolor[HTML]{F3E295}96.9\,\textsubscript{0.3} &  &  & \cellcolor[HTML]{FFFCF3}54.5\,\textsubscript{0.2} & \cellcolor[HTML]{FFFDF6}53.7\,\textsubscript{0.2} \\
  &  & dep \sequential{} & \cellcolor[HTML]{C3D385}97.5\,\textsubscript{0.7} & \cellcolor[HTML]{FFE69B}96.4\,\textsubscript{0.3} &  &  & \cellcolor[HTML]{FFFBED}55.7\,\textsubscript{0.5} & \cellcolor[HTML]{FFFCF1}54.8\,\textsubscript{0.6} \\
  &  & NER\multitask{} & \cellcolor[HTML]{A9CC7C}97.9\,\textsubscript{0.6} & \cellcolor[HTML]{F1E194}96.9\,\textsubscript{0.1} &  &  & \cellcolor[HTML]{FFFCF2}54.6\,\textsubscript{0.5} & \cellcolor[HTML]{FFFDF6}53.8\,\textsubscript{0.3} \\
  &  & NER \sequential{} & \cellcolor[HTML]{C2D384}97.6\,\textsubscript{0.2} & \cellcolor[HTML]{FFE69C}96.4\,\textsubscript{0.5} &  &  & \cellcolor[HTML]{FFFDF4}54.1\,\textsubscript{0.6} & \cellcolor[HTML]{FFFDF7}53.5\,\textsubscript{1.0} \\
 \horizontalspacer
 & mDeBERTa & baseline & \cellcolor[HTML]{89C271}{\color[HTML]{F3F3F3} 98.4} {\textsubscript{0.4}} & \cellcolor[HTML]{FCE498}96.7\,\textsubscript{0.3} &  &  & \cellcolor[HTML]{FFFAE9}56.5\,\textsubscript{0.2} & \cellcolor[HTML]{FFFBEF}55.2\,\textsubscript{1.1} \\
  &  & noise (10\%) & \cellcolor[HTML]{7EBF6D}{\color[HTML]{F3F3F3} 98.5\,\textsubscript{0.6}} & \cellcolor[HTML]{FFE69B}96.5\,\textsubscript{0.9} &  &  & \cellcolor[HTML]{FFFAE8}56.7\,\textsubscript{0.7} & \cellcolor[HTML]{FFFBEE}55.6\,\textsubscript{1.0} \\
  &  & noise (20\%) & \cellcolor[HTML]{52B15E}{\color[HTML]{F3F3F3} 99.2\,\textsubscript{0.1}} & \cellcolor[HTML]{C6D486}97.5\,\textsubscript{0.2} &  \cellcolor[HTML]{BDD182} 97.6 &  & \cellcolor[HTML]{FFFAEA}56.4\,\textsubscript{0.2} & \cellcolor[HTML]{FFFBEE}55.5\,\textsubscript{0.5} \\
  &  & noise (30\%) & \cellcolor[HTML]{66B765}{\color[HTML]{F3F3F3} 98.9\,\textsubscript{0.5}} & \cellcolor[HTML]{E5DE91}97.0\,\textsubscript{0.5} &  &  & \cellcolor[HTML]{FFF9E5}57.6\,\textsubscript{0.3} & \cellcolor[HTML]{FFFAEB}56.2\,\textsubscript{0.5} \\
 \midrule
Nor. dialect & NorBERT & baseline\textsuperscript{1} & \cellcolor[HTML]{43AD58}{\color[HTML]{F3F3F3} 99.4\,\asteriskdev{}\textsubscript{0.0}} & \cellcolor[HTML]{FFEAAD}94.2\,\textsubscript{0.6} &  &  & \cellcolor[HTML]{46AE5A}{\color[HTML]{F3F3F3} 94.5\,\asteriskdev{}\textsubscript{0.6}} & \cellcolor[HTML]{E7DE91}76.8\,\textsubscript{1.1} \\
(dev 90\%) & ScandiBERT & baseline\textsuperscript{1} & \cellcolor[HTML]{34A853}{\color[HTML]{F3F3F3} 99.6\,\asteriskdev{}\textsubscript{0.2}} & \cellcolor[HTML]{FFEDB8}92.8\,\textsubscript{0.6} &  &  & \cellcolor[HTML]{38AA55}{\color[HTML]{F3F3F3} 96.0\,\asteriskdev{}\textsubscript{0.6}} & \cellcolor[HTML]{BFD283}81.2\,\textsubscript{0.6} \\
 & mDeBERTa & baseline\textsuperscript{1} & \cellcolor[HTML]{43AD58}{\color[HTML]{F3F3F3} 99.4\,\asteriskdev{}\textsubscript{0.0}} & \cellcolor[HTML]{FFECB3}93.4\,\textsubscript{0.7} & \cellcolor[HTML]{FFECB3}{ 93.5} &  & \cellcolor[HTML]{34A853}{\color[HTML]{F3F3F3} 96.4\,\asteriskdev{}\textsubscript{0.4}} & \cellcolor[HTML]{ADCD7D}83.2\,\textsubscript{1.0} & \cellcolor[HTML]{B0CE7E}83.7 \\
 \midrule
Nor. dialect & mDeBERTa & EnSID expert & \cellcolor[HTML]{FFE9A6}95.1\,\textsubscript{0.6} & \cellcolor[HTML]{FFEFBF}92.0\,\textsubscript{0.8} &  &  & \cellcolor[HTML]{D6D98B}78.6\,\textsubscript{1.1} & \cellcolor[HTML]{E3DD90}77.2\,\textsubscript{1.6} \\
(dev 90\%) /  &  & NorSID expert\textsuperscript{1,2} & \cellcolor[HTML]{43AD58}{\color[HTML]{F3F3F3} 99.4\,\asteriskdev{}\textsubscript{0.0}} & \cellcolor[HTML]{FFECB3}93.4\,\textsubscript{0.7} &  &  & \cellcolor[HTML]{34A853}{\color[HTML]{F3F3F3} 96.4\,\asteriskdev{}\textsubscript{0.4}} & \cellcolor[HTML]{ADCD7D}83.2\,\textsubscript{1.0} \\
English  &  & assembled & \cellcolor[HTML]{8EC372}{\color[HTML]{F3F3F3} 98.3\,\textsubscript{0.4}} & \cellcolor[HTML]{FFE69B}96.4\,\textsubscript{0.2} & \cellcolor[HTML]{DEDB8E}{ 97.2} &  & \cellcolor[HTML]{8EC372}86.5\,\textsubscript{0.6} & \cellcolor[HTML]{9DC878}84.9\,\textsubscript{0.5} & \cellcolor[HTML]{9FC978}85.6  \\
 \bottomrule
\end{tabular}
}
\caption{\textbf{Intent classification and slot-filling scores for all systems} on the development and test data, and for the runs we submitted to the shared task. 
Results are averaged across three runs, with the exception of the assembled system, which is averaged across nine total combinations of three runs each of both experts. Standard deviations are denoted by subscripts.
Key: 
\textit{Dial}\,=\,dialect identification, 
\textit{dep}\,=\,dependency parsing,
\multitask{}\,=\,multitask learning,
\sequential{}\,=\,intermediate-task training.
\textsuperscript{1}For the models trained on 90\% of the development data, the dev scores are measured on the remaining 10\%.
\textsuperscript{2}The results for this model are already listed in the Norwegian dialect section (mDeBERTa), but repeated here for easier comparison.
}
\label{tab:results-all}
\end{table*}

\begin{table*}[t]
\setlength{\tabcolsep}{3pt}
\centering
\begin{tabular}{@{}llrrr@{\,\,}l@{}r@{\,\,}r@{}}
\toprule
\textbf{Training data} & \textbf{Model} & \multicolumn{2}{@{}l}{\textbf{Loose}} & \multicolumn{2}{@{}l}{\textbf{\llap{U}nlabelled}} & \multicolumn{2}{l}{\textbf{Strict}} \\
\midrule
English & NorBERT & \cellcolor[HTML]{FFFFFF}84.4 & \textsubscript{1.0} & \cellcolor[HTML]{88CFAC}84.4 & \textsubscript{1.3} & \cellcolor[HTML]{7ECBA5}76.5 & \textsubscript{1.2} \\
(train) & ScandiBERT & \cellcolor[HTML]{57BB8A}88.0 & \textsubscript{0.3} & \cellcolor[HTML]{73C79E}88.2 & \textsubscript{0.2} & \cellcolor[HTML]{66C194}80.7 & \textsubscript{0.7} \\
 & mDeBERTa & \cellcolor[HTML]{8FD2B1}86.8 & \textsubscript{0.2} & \cellcolor[HTML]{7AC9A2}87.0 & \textsubscript{0.7} & \cellcolor[HTML]{6CC499}79.7 & \textsubscript{0.4} \\
\horizontalspacer
Norwegian (MT) & NorBERT & \cellcolor[HTML]{FCFEFD}84.4 & \textsubscript{0.5} & \cellcolor[HTML]{FDFEFE}63.4 & \textsubscript{0.7} & \cellcolor[HTML]{FFFFFF}53.9 & \textsubscript{0.3} \\
(train) & ScandiBERT & \cellcolor[HTML]{B6E2CC}86.0 & \textsubscript{0.6} & \cellcolor[HTML]{FFFFFF}62.9 & \textsubscript{0.4} & \cellcolor[HTML]{FCFEFD}54.6 & \textsubscript{0.4} \\
 & mDeBERTa & \cellcolor[HTML]{C0E6D3}85.7 & \textsubscript{0.4} & \cellcolor[HTML]{FEFFFE}63.3 & \textsubscript{1.1} & \cellcolor[HTML]{F8FDFA}55.2 & \textsubscript{1.1} \\
\horizontalspacer
Nor. dialect & NorBERT & \cellcolor[HTML]{E9F6F0}84.9 & \textsubscript{1.0} & \cellcolor[HTML]{66C195}90.5 & \textsubscript{0.2} & \cellcolor[HTML]{7CCAA4}76.8 & \textsubscript{1.1} \\
(dev 90\%) & ScandiBERT & \cellcolor[HTML]{5CBD8E}87.9 & \textsubscript{0.4} & \cellcolor[HTML]{57BB8A}93.1 & \textsubscript{0.3} & \cellcolor[HTML]{57BB8A}83.2 & \textsubscript{1.0} \\
 & mDeBERTa & \cellcolor[HTML]{99D6B8}86.6 & \textsubscript{0.5} & \cellcolor[HTML]{5ABD8C}92.6 & \textsubscript{0.2} & \cellcolor[HTML]{63C093}81.2 & \textsubscript{0.6}\\
\bottomrule
\end{tabular}
\caption{\textbf{Test scores of baseline models on slot filling for \fonescore{} variants: loose, unlabelled, and strict span} (all \fonescore{} scores in~\%). Strict span is the \fonescore{} score we use throughout, where both the span and label must be fully correct, loose \fonescore{} allows for partial matches of the span (if the label is correct), and unlabelled ignores the label (considering only the span overlaps).
The results are averaged over three runs, with standard deviations as subscripts.
}
\label{tab:results-baselines-slots-detailed}
\end{table*}

\begin{table*}[t]
\centering
\setlength{\tabcolsep}{3pt}
\begin{tabular}{@{}llrrrrcrrrr@{}}
\toprule
 & & \multicolumn{4}{c}{\textbf{Intents}} && \multicolumn{4}{c}{\textbf{Slots}} \\
\cmidrule(lr){3-6} \cmidrule(lr){7-11}
\textbf{PLM} & \textbf{Split} & \multicolumn{1}{c}{\boldmath$r$} & \multicolumn{1}{c}{\textit{\textbf{p}}\boldmath$_r$} & \multicolumn{1}{c}{\boldmath$\rho$} & \multicolumn{1}{c}{\textit{\textbf{p}}\boldmath$_\rho$} && \multicolumn{1}{c}{\boldmath$r$} & \multicolumn{1}{c}{\textit{\textbf{p}}\boldmath$_r$} & \multicolumn{1}{c}{\boldmath$\rho$} & \multicolumn{1}{c}{\textit{\textbf{p}}\boldmath$_\rho$} \\
\midrule
mDeBERTa & dev & \cellcolor[HTML]{B5D2ED}--0.51 & \cellcolor[HTML]{CCCCCC}0.09 & \cellcolor[HTML]{A8CAEA}--0.60 & 0.04 && \cellcolor[HTML]{AECEEB}--0.56 & \cellcolor[HTML]{CCCCCC}0.06 & \cellcolor[HTML]{BED7EF}--0.45 & \cellcolor[HTML]{CCCCCC}0.14 \\
 & test & \cellcolor[HTML]{C8DDF1}--0.38 & \cellcolor[HTML]{CCCCCC}0.22 & \cellcolor[HTML]{BFD8EF}--0.44 & \cellcolor[HTML]{CCCCCC}0.15 && \cellcolor[HTML]{C2DAF0}--0.42 & \cellcolor[HTML]{CCCCCC}0.17 & \cellcolor[HTML]{CCE0F2}--0.35 & \cellcolor[HTML]{CCCCCC}0.27 \\
 & dev+test & \cellcolor[HTML]{CBDFF2}--0.36 & \cellcolor[HTML]{CCCCCC}0.08 & \cellcolor[HTML]{BFD8EF}--0.44 & 0.03 && \cellcolor[HTML]{BBD6EE}--0.47 & 0.02 & \cellcolor[HTML]{C6DDF1}--0.39 & \cellcolor[HTML]{CCCCCC}0.06 \\
\horizontalspacer
ScandiBERT & dev & \cellcolor[HTML]{ACCDEB}--0.57 & \cellcolor[HTML]{CCCCCC}0.06 & \cellcolor[HTML]{AECEEB}--0.56 & \cellcolor[HTML]{CCCCCC}0.06 && \cellcolor[HTML]{DCEAF6}--0.24 & \cellcolor[HTML]{CCCCCC}0.44 & \cellcolor[HTML]{CEE1F3}--0.34 & \cellcolor[HTML]{CCCCCC}0.28 \\
 & test & \cellcolor[HTML]{9FC5E7}--0.66 & 0.02 & \cellcolor[HTML]{9DC3E7}--0.68 & 0.02 && \cellcolor[HTML]{FFFCF1}0.14 & \cellcolor[HTML]{CCCCCC}0.66 & \cellcolor[HTML]{FFFEF8}0.07 & \cellcolor[HTML]{CCCCCC}0.83 \\
 & dev+test & \cellcolor[HTML]{B2D0EC}--0.53 & 0.01 & \cellcolor[HTML]{B7D3ED}--0.50 & 0.01 && \cellcolor[HTML]{E7F1F9}--0.16 & \cellcolor[HTML]{CCCCCC}0.45 & \cellcolor[HTML]{E3EEF8}--0.19 & \cellcolor[HTML]{CCCCCC}0.36 \\
\horizontalspacer
NorBERT & dev & \cellcolor[HTML]{9AC2E6}--0.70 & 0.01 & \cellcolor[HTML]{A4C8E8}--0.63 & 0.03 && \cellcolor[HTML]{E7F1F9}--0.16 & \cellcolor[HTML]{CCCCCC}0.63 & \cellcolor[HTML]{D3E4F4}--0.30 & \cellcolor[HTML]{CCCCCC}0.34 \\
 & test & \cellcolor[HTML]{88B7E2}--0.82 & 0.00 & \cellcolor[HTML]{8AB8E2}--0.81 & 0.00 && \cellcolor[HTML]{FAFCFD}--0.03 & \cellcolor[HTML]{CCCCCC}0.93 & \cellcolor[HTML]{F2F7FB}--0.09 & \cellcolor[HTML]{CCCCCC}0.79 \\
 & dev+test & \cellcolor[HTML]{B8D4ED}--0.49 & 0.01 & \cellcolor[HTML]{AECEEB}--0.56 & 0.00 && \cellcolor[HTML]{E5EFF8}--0.18 & \cellcolor[HTML]{CCCCCC}0.40 & \cellcolor[HTML]{D8E7F5}--0.27 & \cellcolor[HTML]{CCCCCC}0.20
\\
\bottomrule
\end{tabular}
\caption{\textbf{Correlations between the split word ratio difference and SID performance for the noising experiments:} Pearson's~\textit{r} and Spearman's~$\rho$ with corresponding \textit{p}-values (\textit{p}-values $\geq$0.05 have a grey background). Each dev or test row is based on twelve observations (four noise levels à three initializations).}
\label{tab:noise-correlations}
\end{table*}

\begin{table*}
\setlength{\tabcolsep}{3pt}
\centering
\adjustbox{max width=\linewidth}{%
\begin{tabular}{@{}l@{\,}lr@{\,}r@{\,\,}rr@{\,}r@{\,\,}rr@{\,}r@{\,\,}rr@{\,}r@{\,\,}rr@{\,}rr@{\,}rr@{\,}rr@{\,}r@{}}
\toprule
&& \multicolumn{6}{c}{\textbf{Intents}} & \multicolumn{6}{c}{\textbf{Slots}} & \multicolumn{8}{c}{\textbf{Aux.\ task performance (dev)}} \\
\cmidrule(lr){3-8} \cmidrule(lr){9-14} \cmidrule(lr){15-22}
\multicolumn{2}{l}{\textbf{Task}} & \textbf{Dev} &  & \boldmath$\Delta_{dev}$ & \textbf{Test} &  & \boldmath$\Delta_{test}$ & \textbf{Dev} &  & \boldmath$\Delta_{dev}$ & \textbf{Test} &  & \boldmath$\Delta_{test}$ & {\textbf{Dial}} &  & {\textbf{POS}} &  & \textbf{Dep} &  & \textbf{NER} &  \\
\midrule
\multicolumn{20}{l}{\textit{English SID training data}} \\
\multicolumn{2}{@{}l}{Baseline}& 96.4 & \textsubscript{0.5} &  & 94.8 & \textsubscript{0.8} &  & 81.3 & \textsubscript{0.3} &  & 80.7 & \textsubscript{0.7} &  &  &  &  &  &  &  &  &  \\
\horizontalspacer
Dial& \multitask{} & 84.3 & \textsubscript{2.7} & \cellcolor[HTML]{E67C73}--12.1 & 83.8 & \textsubscript{3.2} & \cellcolor[HTML]{E67C73}--11.0 & 75.8 & \textsubscript{1.8} & \cellcolor[HTML]{F2BDB9}--5.5 & 75.8 & \textsubscript{1.2} & \cellcolor[HTML]{F3C4C0}--4.9 & {75.9} & {\textsubscript{3.5}} &  &  &  &  & &  \\
& \sequential{} & 95.8 & \textsubscript{0.7} & \cellcolor[HTML]{FDF7F7}--0.6 & 94.0 & \textsubscript{1.7} & \cellcolor[HTML]{FDF4F4}--0.8 & 79.7 & \textsubscript{1.3} & \cellcolor[HTML]{FBEBEA}--1.6 & 79.2 & \textsubscript{0.9} & \cellcolor[HTML]{FBEDEB}--1.5 & {80.0} & {\textsubscript{0.3}} &  &  &  &  &  &  \\
\horizontalspacer
POS& \multitask{} & 96.0 & \textsubscript{0.5} & \cellcolor[HTML]{FEFAFA}--0.4 & 94.9 & \textsubscript{0.3} & \cellcolor[HTML]{FFFFFF}+0.0 & 81.6 & \textsubscript{0.2} & \cellcolor[HTML]{F7FCF9}+0.3 & 81.1 & \textsubscript{0.3} & \cellcolor[HTML]{F1F9F5}+0.4 &  &  & {79.5} & {\textsubscript{0.6}} &  &  &  &  \\
& \sequential{} & 96.3 & \textsubscript{0.2} & \cellcolor[HTML]{FEFDFD}--0.1 & 94.7 & \textsubscript{0.2} & \cellcolor[HTML]{FEFDFC}--0.2 & 82.3 & \textsubscript{0.8} & \cellcolor[HTML]{DDF2E8}+1.0 & 82.2 & \textsubscript{1.1} & \cellcolor[HTML]{CFECDE}+1.5 &  &  & {92.1} & {\textsubscript{0.0}} &  &  &  &  \\
\horizontalspacer
Dep & \multitask{} & 94.7 & \textsubscript{1.2} & \cellcolor[HTML]{FBEAE8}--1.8 & 93.5 & \textsubscript{0.6} & \cellcolor[HTML]{FCEFEE}--1.3 & 82.0 & \textsubscript{0.4} & \cellcolor[HTML]{E8F6EF}+0.7 & 81.5 & \textsubscript{0.2} & \cellcolor[HTML]{E6F5ED}+0.8 &  &  &  &  & {46.6} & {\textsubscript{0.8}} &  &  \\
& \sequential{} & 96.7 & \textsubscript{1.0} & \cellcolor[HTML]{F6FCF9}+0.3 & 94.9 & \textsubscript{1.2} & \cellcolor[HTML]{FEFFFE}+0.1 & 82.5 & \textsubscript{0.9} & \cellcolor[HTML]{D7EFE3}+1.2 & 81.8 & \textsubscript{0.7} & \cellcolor[HTML]{DBF1E6}+1.1 &  &  &  &  & {67.8} & {\textsubscript{0.6}} &  &  \\
\horizontalspacer
NER & \multitask{} & 97.2 & \textsubscript{0.9} & \cellcolor[HTML]{E5F5ED}+0.8 & 95.3 & \textsubscript{1.0} & \cellcolor[HTML]{EFF9F4}+0.5 & 80.9 & \textsubscript{0.9} & \cellcolor[HTML]{FEFAF9}--0.4 & 80.6 & \textsubscript{1.0} & \cellcolor[HTML]{FEFDFD}--0.1 &  &  &  &  &  &  & {93.2} & {\textsubscript{0.2}} \\
& \sequential{} & 96.8 & \textsubscript{0.4} & \cellcolor[HTML]{F2FAF6}+0.4 & 95.0 & \textsubscript{0.4} & \cellcolor[HTML]{FBFEFD}+0.1 & 81.3 & \textsubscript{1.1} & \cellcolor[HTML]{FEFFFF}+0.0 & 81.1 & \textsubscript{0.9} & \cellcolor[HTML]{F1FAF6}+0.4 &  &  &  &  &  &  & {93.0} & {\textsubscript{0.1}} \\
\midrule
\multicolumn{20}{l}{\textit{Machine-translated Norwegian SID training data}} \\
\multicolumn{2}{@{}l}{Baseline}& 97.6 & \textsubscript{0.0} &  & 96.3 & \textsubscript{0.1} &  & 55.5 & \textsubscript{0.4} &  & 54.6 & \textsubscript{0.4} &  &  &  &  &  &  &  &  &  \\
\horizontalspacer
Dial & \multitask{} & 89.8 & \textsubscript{1.4} & \cellcolor[HTML]{EDA29C}--7.8 & 89.2 & \textsubscript{1.4} & \cellcolor[HTML]{EEAAA4}--7.1 & 53.0 & \textsubscript{0.1} & \cellcolor[HTML]{F9E0DE}--2.6 & 51.7 & \textsubscript{0.1} & \cellcolor[HTML]{F8DBD9}--2.9 & {77.1} & {\textsubscript{1.2}} &  &  &  &  &  &  \\
& \sequential{} & 96.1 & \textsubscript{1.0} & \cellcolor[HTML]{FBECEB}--1.5 & 95.2 & \textsubscript{1.0} & \cellcolor[HTML]{FCF2F1}--1.1 & 54.4 & \textsubscript{0.4} & \cellcolor[HTML]{FCF1F0}--1.2 & 53.7 & \textsubscript{0.5} & \cellcolor[HTML]{FCF4F3}--0.9 & {79.7} & {\textsubscript{0.3}} &  &  &  &  &  &  \\
\horizontalspacer
POS & \multitask{} & 97.9 & \textsubscript{0.3} & \cellcolor[HTML]{F4FBF8}+0.3 & 96.8 & \textsubscript{0.4} & \cellcolor[HTML]{F1FAF5}+0.4 & 54.0 & \textsubscript{0.5} & \cellcolor[HTML]{FBEDEB}--1.5 & 53.7 & \textsubscript{0.6} & \cellcolor[HTML]{FCF4F3}--0.9 &  &  & {70.3} & {\textsubscript{1.4}} &  &  &  &  \\
& \sequential{} & 97.8 & \textsubscript{0.5} & \cellcolor[HTML]{FAFDFC}+0.2 & 96.7 & \textsubscript{0.4} & \cellcolor[HTML]{F4FBF7}+0.3 & 55.5 & \textsubscript{0.4} & \cellcolor[HTML]{FFFFFF}+0.0 & 54.4 & \textsubscript{0.8} & \cellcolor[HTML]{FEFCFC}--0.2 &  &  & {92.1} & {\textsubscript{0.1}} &  &  &  &  \\
\horizontalspacer
Dep & \multitask{} & 98.0 & \textsubscript{0.4} & \cellcolor[HTML]{F4FBF7}+0.4 & 96.9 & \textsubscript{0.3} & \cellcolor[HTML]{EDF8F3}+0.5 & 54.5 & \textsubscript{0.2} & \cellcolor[HTML]{FCF3F2}--1.0 & 53.7 & \textsubscript{0.2} & \cellcolor[HTML]{FCF4F3}--0.9 &  &  &  &  & {41.1} & {\textsubscript{0.9}} &  &  \\
& \sequential{} & 97.5 & \textsubscript{0.7} & \cellcolor[HTML]{FEFEFE}--0.1 & 96.4 & \textsubscript{0.3} & \cellcolor[HTML]{FBFEFC}+0.1 & 55.7 & \textsubscript{0.5} & \cellcolor[HTML]{FAFDFC}+0.2 & 54.8 & \textsubscript{0.6} & \cellcolor[HTML]{F9FDFB}+0.2 &  &  &  &  & {67.8} & {\textsubscript{0.6}} &  &  \\
\horizontalspacer
NER & \multitask{} & 97.9 & \textsubscript{0.6} & \cellcolor[HTML]{F5FBF8}+0.3 & 96.9 & \textsubscript{0.1} & \cellcolor[HTML]{ECF8F2}+0.6 & 54.6 & \textsubscript{0.5} & \cellcolor[HTML]{FCF3F3}--0.9 & 53.8 & \textsubscript{0.3} & \cellcolor[HTML]{FDF5F5}--0.8 &  &  &  &  &  &  & {92.3} & {\textsubscript{0.3}} \\
& \sequential{} & 97.6 & \textsubscript{0.2} & \cellcolor[HTML]{FEFEFE}--0.0 & 96.4 & \textsubscript{0.5} & \cellcolor[HTML]{FDFFFE}+0.1 & 54.1 & \textsubscript{0.6} & \cellcolor[HTML]{FBEEEC}--1.4 & 53.5 & \textsubscript{1.0} & \cellcolor[HTML]{FCF2F1}--1.1 &  &  &  &  &  &  & {93.0} & {\textsubscript{0.1}} \\
 \bottomrule
\end{tabular}
}
\caption{\textbf{Performance of the models trained on auxiliary task data} on the SID data (development and test) and the auxiliary tasks (development sets).
Scores are averaged over three runs (standard deviations in subscript numbers) and in~\% -- intent classification: accuracy, slot filling: span~\fonescore{}, dialect classification (``dial''): accuracy, POS tagging: accuracy, dependency parsing (``dep''): labelled attachment score, NER: span~\fonescore{}.
The $\Delta$ columns show the differences to the respective baseline.
Joint multi-task learning is denoted by a~\multitask{}, and intermediate-task training by a~\sequential{}.
}
\label{tab:auxtask-details}
\end{table*}

\begin{table*}
\setlength{\tabcolsep}{1pt}
\centering
\adjustbox{max width=\linewidth}{%
\begin{tabular}{@{}lr@{\skiptiny}r@{\skipsmall}r@{\skipmid}r@{\skiptiny}r@{\skipsmall}r@{\skipmid}r@{\skiptiny}r@{\skipsmall}r@{\skipmid}r@{\skiptiny}r@{\skipsmall}r@{\skipmid}r@{\skiptiny}r@{\skipsmall}r@{\skiplarge}r@{\skiptiny}r@{\skipsmall}r@{\skipmid}r@{\skiptiny}r@{\skipsmall}r@{\skipmid}r@{\skiptiny}r@{\skipsmall}r@{\skipmid}r@{\skiptiny}r@{\skipsmall}r@{\skipmid}r@{\skiptiny}r@{\skipsmall}r@{}}
\multicolumn{31}{c}{\textit{{Trained on auxiliary tasks and English SID data}}} \\
\toprule
& \multicolumn{15}{c}{\textbf{Intents (acc., \%)}} & \multicolumn{15}{c}{\textbf{Slots (span \fonescore, \%)}} \\
\cmidrule(lr){2-16} \cmidrule(lr){17-31}
\textbf{Aux} & {\textbf{B}} & & \multicolumn{1}{c}{\boldmath$\Delta_B$} & {\textbf{N}} & & \multicolumn{1}{c}{\boldmath$\Delta_N$} & {\textbf{T}} & & \multicolumn{1}{c}{\boldmath$\Delta_T$} & {\textbf{W}} & & \multicolumn{1}{c}{\boldmath$\Delta_W$} & {\textbf{all}} & & \multicolumn{1}{c}{\boldmath$\Delta_{all}$} & {\textbf{B}} & & \multicolumn{1}{c}{\boldmath$\Delta_B$} & {\textbf{N}} & & \multicolumn{1}{c}{\boldmath$\Delta_N$} & {\textbf{T}} & & \multicolumn{1}{c}{\boldmath$\Delta_T$} & {\textbf{W}} & & \multicolumn{1}{c}{\boldmath$\Delta_W$} & {\textbf{all}} & & \multicolumn{1}{c}{\boldmath$\Delta_{all}$} \\ 
 \midrule
\textit{none} & 96.3 & \textsubscript{0.7} & \multicolumn{1}{l}{} & 93.3 & \textsubscript{1.5} & \multicolumn{1}{l}{} & 94.0 & \textsubscript{0.9} & \multicolumn{1}{l}{} & 95.6 & \textsubscript{0.5} & \multicolumn{1}{l}{} & 94.8 & \textsubscript{0.8} & \multicolumn{1}{l}{} & 83.7 & \textsubscript{1.0} & \multicolumn{1}{l}{} & 75.7 & \textsubscript{0.8} & \multicolumn{1}{l}{} & 79.3 & \textsubscript{0.7} & \multicolumn{1}{l}{} & 82.8 & \textsubscript{0.8} & \multicolumn{1}{l}{} & 80.7 & \textsubscript{0.7} & \multicolumn{1}{l}{} \\
\midrule
\midrule
& \multicolumn{3}{@{}c@{}}{\cellcolor[HTML]{EEF4FA}\textit{9.2\%}} & \multicolumn{3}{@{}c@{}}{\cellcolor[HTML]{DCE9F5}\textit{18.5\%}} & \multicolumn{3}{@{}c@{}}{\cellcolor[HTML]{CBDFF0}\textit{26.9\%}} & \multicolumn{3}{@{}c@{}}{\cellcolor[HTML]{A7C8E6}\textit{45.4\%}} &  &  &  & \multicolumn{3}{@{}c@{}}{\cellcolor[HTML]{EEF4FA}\textit{9.2\%}} & \multicolumn{3}{@{}c@{}}{\cellcolor[HTML]{DCE9F5}\textit{18.5\%}} & \multicolumn{3}{@{}c@{}}{\cellcolor[HTML]{CBDFF0}\textit{26.9\%}} & \multicolumn{3}{@{}c@{}}{\cellcolor[HTML]{A7C8E6}\textit{45.4\%}} &  &  & \\
\midrule
Dial \multitask & 87.3 & \textsubscript{0.6} & \cellcolor[HTML]{F3C4C0}--9.1 & 72.8 & \textsubscript{6.1} & \cellcolor[HTML]{E67C73}--20.4 & 81.0 & \textsubscript{5.0} & \cellcolor[HTML]{EFABA6}--13.0 & 89.2 & \textsubscript{1.7} & \cellcolor[HTML]{F7D5D2}--6.5 & 83.8 & \textsubscript{3.2} & \cellcolor[HTML]{F1B8B3}--11.0 & 82.7 & \textsubscript{0.9} & \cellcolor[HTML]{FDF8F8}--1.0 & 70.0 & \textsubscript{2.2} & \cellcolor[HTML]{F7DAD7}--5.7 & 70.1 & \textsubscript{1.8} & \cellcolor[HTML]{F3C3BF}--9.3 & 79.9 & \textsubscript{0.8} & \cellcolor[HTML]{FBECEA}--2.9 & 75.8 & \textsubscript{1.2} & \cellcolor[HTML]{F8DFDD}--4.9 \\
Dial \sequential & 96.5 & \textsubscript{0.9} & \cellcolor[HTML]{F5FBF8}+0.1 & 92.0 & \textsubscript{3.5} & \cellcolor[HTML]{FDF6F6}--1.3 & 92.0 & \textsubscript{1.6} & \cellcolor[HTML]{FCF2F1}--2.0 & 95.5 & \textsubscript{1.3} & \cellcolor[HTML]{FEFDFD}--0.2 & 94.0 & \textsubscript{1.7} & \cellcolor[HTML]{FDF9F9}--0.8 & 81.9 & \textsubscript{0.8} & \cellcolor[HTML]{FCF3F2}--1.9 & 75.4 & \textsubscript{1.8} & \cellcolor[HTML]{FEFCFC}--0.4 & 76.2 & \textsubscript{1.5} & \cellcolor[HTML]{FBEAE9}--3.2 & 82.0 & \textsubscript{0.2} & \cellcolor[HTML]{FDF9F8}--0.9 & 79.2 & \textsubscript{0.9} & \cellcolor[HTML]{FDF5F4}--1.5 \\
\midrule
\midrule
 & \multicolumn{3}{@{}c@{}}{\textit{0.0\%}} & \multicolumn{3}{@{}c@{}}{\cellcolor[HTML]{C9DDEF}\textit{28.1\%}} & \multicolumn{3}{@{}c@{}}{\cellcolor[HTML]{F1F6FB}\textit{7.6\%}} & \multicolumn{3}{@{}c@{}}{\cellcolor[HTML]{BED6EC}\textit{33.7\%}}  &  &  &  & \multicolumn{3}{@{}c@{}}{\textit{0.0\%}} & \multicolumn{3}{@{}c@{}}{\cellcolor[HTML]{C9DDEF}\textit{28.1\%}} & \multicolumn{3}{@{}c@{}}{\cellcolor[HTML]{F1F6FB}\textit{7.6\%}} & \multicolumn{3}{@{}c@{}}{\cellcolor[HTML]{BED6EC}\textit{33.7\%}} \\
\midrule
POS \multitask & 95.9 & \textsubscript{0.9} & \cellcolor[HTML]{FEFCFB}--0.5 & 93.1 & \textsubscript{0.6} & \cellcolor[HTML]{FEFDFD}--0.2 & 94.0 & \textsubscript{0.9} & \cellcolor[HTML]{FFFFFF}+0.0 & 95.9 & \textsubscript{0.4} & \cellcolor[HTML]{EFF9F4}+0.2 & 94.9 & \textsubscript{0.3} & \cellcolor[HTML]{FEFFFE}+0.0 & 83.7 & \textsubscript{0.5} & \cellcolor[HTML]{FEFEFE}--0.1 & 77.0 & \textsubscript{0.2} & \cellcolor[HTML]{9AD6B9}+1.3 & 80.5 & \textsubscript{0.8} & \cellcolor[HTML]{A5DBC0}+1.2 & 82.6 & \textsubscript{0.3} & \cellcolor[HTML]{FEFDFD}--0.2 & 81.1 & \textsubscript{0.3} & \cellcolor[HTML]{DDF2E7}+0.4 \\
POS \sequential & 96.2 & \textsubscript{0.4} & \cellcolor[HTML]{FEFEFE}--0.1 & 92.8 & \textsubscript{0.6} & \cellcolor[HTML]{FEFCFB}--0.5 & 93.7 & \textsubscript{0.2} & \cellcolor[HTML]{FEFCFC}--0.4 & 95.7 & \textsubscript{0.5} & \cellcolor[HTML]{FBFEFD}+0.1 & 94.7 & \textsubscript{0.2} & \cellcolor[HTML]{FEFDFD}--0.2 & 85.1 & \textsubscript{0.7} & \cellcolor[HTML]{99D6B8}+1.3 & 77.3 & \textsubscript{1.7} & \cellcolor[HTML]{81CCA7}+1.6 & 81.5 & \textsubscript{1.2} & \cellcolor[HTML]{57BB8A}+2.2 & 83.8 & \textsubscript{0.9} & \cellcolor[HTML]{B2E0CA}+1.0 & 82.2 & \textsubscript{1.1} & \cellcolor[HTML]{8ED2B1}+1.5 \\
Dep \multitask & 95.0 & \textsubscript{0.7} & \cellcolor[HTML]{FDF6F5}--1.3 & 91.6 & \textsubscript{0.8} & \cellcolor[HTML]{FDF4F3}--1.6 & 91.2 & \textsubscript{1.7} & \cellcolor[HTML]{FBECEB}--2.8 & 95.4 & \textsubscript{0.1} & \cellcolor[HTML]{FEFDFD}--0.3 & 93.5 & \textsubscript{0.6} & \cellcolor[HTML]{FDF6F6}--1.3 & 83.8 & \textsubscript{0.6} & \cellcolor[HTML]{FCFEFD}+0.0 & 77.7 & \textsubscript{0.8} & \cellcolor[HTML]{63C093}+2.0 & 80.4 & \textsubscript{0.3} & \cellcolor[HTML]{ADDEC6}+1.1 & 83.1 & \textsubscript{0.2} & \cellcolor[HTML]{EAF7F1}+0.3 & 81.5 & \textsubscript{0.2} & \cellcolor[HTML]{C3E7D6}+0.8 \\
Dep \sequential & 95.9 & \textsubscript{1.2} & \cellcolor[HTML]{FEFCFB}--0.5 & 93.0 & \textsubscript{1.3} & \cellcolor[HTML]{FEFDFD}--0.2 & 94.3 & \textsubscript{1.9} & \cellcolor[HTML]{EBF7F1}+0.3 & 95.8 & \textsubscript{0.8} & \cellcolor[HTML]{F5FBF8}+0.1 & 94.9 & \textsubscript{1.2} & \cellcolor[HTML]{FBFEFC}+0.1 & 84.0 & \textsubscript{0.8} & \cellcolor[HTML]{EBF7F1}+0.3 & 77.5 & \textsubscript{1.9} & \cellcolor[HTML]{74C79E}+1.8 & 80.3 & \textsubscript{1.2} & \cellcolor[HTML]{B6E2CC}+0.9 & 83.9 & \textsubscript{0.4} & \cellcolor[HTML]{AEDFC7}+1.0 & 81.8 & \textsubscript{0.7} & \cellcolor[HTML]{ABDDC5}+1.1 \\
\midrule
\midrule
& \multicolumn{3}{c}{\cellcolor[HTML]{3D85C6}{\color[HTML]{FFFFFF} \textit{100.0\%}}} & \multicolumn{3}{@{}c@{}}{\textit{0.0\%}}& \multicolumn{3}{@{}c@{}}{\textit{0.0\%}}& \multicolumn{3}{@{}c@{}}{\textit{0.0\%}} &&&& \multicolumn{3}{c}{\cellcolor[HTML]{3D85C6}{\color[HTML]{FFFFFF} \textit{100.0\%}}} & \multicolumn{3}{@{}c@{}}{\textit{0.0\%}}& \multicolumn{3}{@{}c@{}}{\textit{0.0\%}}& \multicolumn{3}{@{}c@{}}{\textit{0.0\%}}\\
\midrule
NER \multitask & 96.4 & \textsubscript{0.7} & \cellcolor[HTML]{FAFDFC}+0.1 & 94.1 & \textsubscript{1.4} & \cellcolor[HTML]{C1E6D4}+0.8 & 95.0 & \textsubscript{1.5} & \cellcolor[HTML]{B7E2CD}+0.9 & 95.8 & \textsubscript{0.6} & \cellcolor[HTML]{F0F9F5}+0.2 & 95.3 & \textsubscript{1.0} & \cellcolor[HTML]{D8F0E4}+0.5 & 83.9 & \textsubscript{2.2} & \cellcolor[HTML]{F1FAF6}+0.2 & 76.2 & \textsubscript{1.0} & \cellcolor[HTML]{D8EFE4}+0.5 & 79.1 & \textsubscript{1.0} & \cellcolor[HTML]{FEFDFD}--0.2 & 82.5 & \textsubscript{0.7} & \cellcolor[HTML]{FEFCFC}--0.4 & 80.6 & \textsubscript{1.0} & \cellcolor[HTML]{FEFEFE}--0.1 \\
NER \sequential & 96.3 & \textsubscript{0.2} & \cellcolor[HTML]{FEFEFE}--0.1 & 93.6 & \textsubscript{0.9} & \cellcolor[HTML]{E3F4EC}+0.4 & 94.5 & \textsubscript{0.7} & \cellcolor[HTML]{D8EFE4}+0.5 & 95.5 & \textsubscript{0.2} & \cellcolor[HTML]{FEFDFD}--0.2 & 95.0 & \textsubscript{0.4} & \cellcolor[HTML]{F6FCF9}+0.1 & 84.5 & \textsubscript{2.0} & \cellcolor[HTML]{C7E9D8}+0.7 & 75.9 & \textsubscript{0.4} & \cellcolor[HTML]{F0F9F5}+0.2 & 79.3 & \textsubscript{1.6} & \cellcolor[HTML]{FFFFFF}+0.0 & 83.5 & \textsubscript{0.4} & \cellcolor[HTML]{CBEADB}+0.7 & 81.1 & \textsubscript{0.9} & \cellcolor[HTML]{DFF2E9}+0.4\\
\bottomrule
\\
\\
\multicolumn{31}{c}{\textit{{Trained on auxiliary tasks and machine-translated Norwegian SID data}}} \\
\toprule
& \multicolumn{15}{c}{\textbf{Intents (acc., \%)}} & \multicolumn{15}{c}{\textbf{Slots (span \fonescore, \%)}} \\
\cmidrule(lr){2-16} \cmidrule(lr){17-31}
\textbf{Aux} & {\textbf{B}} & & \multicolumn{1}{c}{\boldmath$\Delta_B$} & {\textbf{N}} & & \multicolumn{1}{c}{\boldmath$\Delta_N$} & {\textbf{T}} & & \multicolumn{1}{c}{\boldmath$\Delta_T$} & {\textbf{W}} & & \multicolumn{1}{c}{\boldmath$\Delta_W$} & {\textbf{all}} & & \multicolumn{1}{c}{\boldmath$\Delta_{all}$} & {\textbf{B}} & & \multicolumn{1}{c}{\boldmath$\Delta_B$} & {\textbf{N}} & & \multicolumn{1}{c}{\boldmath$\Delta_N$} & {\textbf{T}} & & \multicolumn{1}{c}{\boldmath$\Delta_T$} & {\textbf{W}} & & \multicolumn{1}{c}{\boldmath$\Delta_W$} & {\textbf{all}} & & \multicolumn{1}{c}{\boldmath$\Delta_{all}$} \\ 
 \midrule
\textit{none} & 97.4 & \textsubscript{0.0} &  & 94.9 & \textsubscript{0.1} &  & 96.9 & \textsubscript{0.5} &  & 96.3 & \textsubscript{0.2} &  & 96.3 & \textsubscript{0.1} &  & 58.7 & \textsubscript{0.3} &  & 50.9 & \textsubscript{1.1} &  & 54.6 & \textsubscript{0.9} &  & 55.2 & \textsubscript{0.4} &  & 54.6 & \textsubscript{0.4} &  \\
\midrule
\midrule
& \multicolumn{3}{@{}c@{}}{\cellcolor[HTML]{EEF4FA}\textit{9.2\%}} & \multicolumn{3}{@{}c@{}}{\cellcolor[HTML]{DCE9F5}\textit{18.5\%}} & \multicolumn{3}{@{}c@{}}{\cellcolor[HTML]{CBDFF0}\textit{26.9\%}} & \multicolumn{3}{@{}c@{}}{\cellcolor[HTML]{A7C8E6}\textit{45.4\%}} &  &  &  & \multicolumn{3}{@{}c@{}}{\cellcolor[HTML]{EEF4FA}\textit{9.2\%}} & \multicolumn{3}{@{}c@{}}{\cellcolor[HTML]{DCE9F5}\textit{18.5\%}} & \multicolumn{3}{@{}c@{}}{\cellcolor[HTML]{CBDFF0}\textit{26.9\%}} & \multicolumn{3}{@{}c@{}}{\cellcolor[HTML]{A7C8E6}\textit{45.4\%}} &  &  & \\
\midrule
Dial \multitask & 95.9 & \textsubscript{0.8} & {\cellcolor[HTML]{FDF5F4}--1.5} & 80.1 & \textsubscript{4.1} & {\cellcolor[HTML]{EC9F99}--14.8} & 86.4 & \textsubscript{2.6} & {\cellcolor[HTML]{F2BBB6}--10.5} & 93.3 & \textsubscript{0.6} & {\cellcolor[HTML]{FBEBEA}--3.0} & 89.2 & \textsubscript{1.4} & {\cellcolor[HTML]{F6D1CE}--7.1} & 57.3 & \textsubscript{0.8} & {\cellcolor[HTML]{FDF5F5}--1.4} & 46.9 & \textsubscript{0.7} & {\cellcolor[HTML]{FAE4E3}--4.1} & 50.3 & \textsubscript{0.5} & {\cellcolor[HTML]{F9E3E1}--4.4} & 53.3 & \textsubscript{0.2} & {\cellcolor[HTML]{FCF2F2}--1.9} & 51.7 & \textsubscript{0.1} & {\cellcolor[HTML]{FBECEA}--2.9} \\
Dial \sequential & 97.6 & \textsubscript{0.0} & {\cellcolor[HTML]{F0F9F5}+0.2} & 92.7 & \textsubscript{0.9} & {\cellcolor[HTML]{FCF0EF}--2.3} & 95.5 & \textsubscript{0.9} & {\cellcolor[HTML]{FDF5F5}--1.4} & 95.7 & \textsubscript{1.4} & {\cellcolor[HTML]{FEFAFA}--0.6} & 95.2 & \textsubscript{1.0} & {\cellcolor[HTML]{FDF8F7}--1.1} & 58.8 & \textsubscript{0.9} & {\cellcolor[HTML]{F5FBF8}+0.1} & 51.0 & \textsubscript{0.2} & {\cellcolor[HTML]{FCFEFD}+0.0} & 52.6 & \textsubscript{0.8} & {\cellcolor[HTML]{FCF2F1}--2.0} & 54.4 & \textsubscript{0.7} & {\cellcolor[HTML]{FEF9F9}--0.8} & 53.7 & \textsubscript{0.5} & {\cellcolor[HTML]{FDF9F8}--0.9} \\
\midrule
\midrule
 & \multicolumn{3}{@{}c@{}}{\textit{0.0\%}} & \multicolumn{3}{@{}c@{}}{\cellcolor[HTML]{C9DDEF}\textit{28.1\%}} & \multicolumn{3}{@{}c@{}}{\cellcolor[HTML]{F1F6FB}\textit{7.6\%}} & \multicolumn{3}{@{}c@{}}{\cellcolor[HTML]{BED6EC}\textit{33.7\%}}  &  &  &  & \multicolumn{3}{@{}c@{}}{\textit{0.0\%}} & \multicolumn{3}{@{}c@{}}{\cellcolor[HTML]{C9DDEF}\textit{28.1\%}} & \multicolumn{3}{@{}c@{}}{\cellcolor[HTML]{F1F6FB}\textit{7.6\%}} & \multicolumn{3}{@{}c@{}}{\cellcolor[HTML]{BED6EC}\textit{33.7\%}} \\
\midrule
POS \multitask & 97.6 & \textsubscript{0.4} & {\cellcolor[HTML]{F0F9F5}+0.2} & 95.6 & \textsubscript{0.6} & {\cellcolor[HTML]{CEEBDD}+0.6} & 97.2 & \textsubscript{0.2} & {\cellcolor[HTML]{EBF7F1}+0.3} & 96.8 & \textsubscript{0.5} & {\cellcolor[HTML]{D7EFE3}+0.5} & 96.8 & \textsubscript{0.4} & {\cellcolor[HTML]{DDF2E8}+0.4} & 57.7 & \textsubscript{0.7} & {\cellcolor[HTML]{FDF8F8}--1.0} & 50.6 & \textsubscript{0.2} & {\cellcolor[HTML]{FEFCFC}--0.4} & 53.9 & \textsubscript{1.3} & {\cellcolor[HTML]{FEFAF9}--0.8} & 54.0 & \textsubscript{0.5} & {\cellcolor[HTML]{FDF7F6}--1.2} & 53.7 & \textsubscript{0.6} & {\cellcolor[HTML]{FDF9F8}--0.9} \\
POS \sequential & 97.5 & \textsubscript{0.4} & {\cellcolor[HTML]{F5FBF8}+0.1} & 95.2 & \textsubscript{0.7} & {\cellcolor[HTML]{EBF7F1}+0.3} & 97.3 & \textsubscript{0.3} & {\cellcolor[HTML]{DEF2E8}+0.4} & 96.7 & \textsubscript{0.4} & {\cellcolor[HTML]{E2F4EB}+0.4} & 96.7 & \textsubscript{0.4} & {\cellcolor[HTML]{E5F5ED}+0.3} & 58.2 & \textsubscript{0.8} & {\cellcolor[HTML]{FEFCFB}--0.4} & 51.3 & \textsubscript{0.6} & {\cellcolor[HTML]{E2F3EB}+0.4} & 54.2 & \textsubscript{0.6} & {\cellcolor[HTML]{FEFCFB}--0.4} & 54.9 & \textsubscript{1.1} & {\cellcolor[HTML]{FEFDFD}--0.3} & 54.4 & \textsubscript{0.8} & {\cellcolor[HTML]{FEFDFD}--0.2} \\
Dep \multitask & 97.1 & \textsubscript{0.1} & {\cellcolor[HTML]{FEFDFD}--0.3} & 95.8 & \textsubscript{0.6} & {\cellcolor[HTML]{BEE5D2}+0.8} & 97.2 & \textsubscript{0.4} & {\cellcolor[HTML]{E3F4EC}+0.4} & 97.0 & \textsubscript{0.1} & {\cellcolor[HTML]{C8E9D9}+0.7} & 96.9 & \textsubscript{0.3} & {\cellcolor[HTML]{D5EEE2}+0.5} & 57.9 & \textsubscript{0.6} & {\cellcolor[HTML]{FEF9F9}--0.8} & 50.8 & \textsubscript{0.1} & {\cellcolor[HTML]{FEFEFD}--0.2} & 53.9 & \textsubscript{0.1} & {\cellcolor[HTML]{FEFAFA}--0.7} & 53.9 & \textsubscript{0.3} & {\cellcolor[HTML]{FDF6F6}--1.3} & 53.7 & \textsubscript{0.2} & {\cellcolor[HTML]{FDF9F8}--0.9} \\
Dep \sequential & 97.5 & \textsubscript{0.3} & {\cellcolor[HTML]{FAFDFC}+0.1} & 95.0 & \textsubscript{0.5} & {\cellcolor[HTML]{F8FCFA}+0.1} & 97.1 & \textsubscript{0.2} & {\cellcolor[HTML]{EEF8F3}+0.2} & 96.4 & \textsubscript{0.5} & {\cellcolor[HTML]{F7FCFA}+0.1} & 96.4 & \textsubscript{0.3} & {\cellcolor[HTML]{F5FBF8}+0.1} & 59.0 & \textsubscript{0.8} & {\cellcolor[HTML]{E6F5EE}+0.3} & 52.0 & \textsubscript{0.4} & {\cellcolor[HTML]{B0DFC8}+1.0} & 54.9 & \textsubscript{0.8} & {\cellcolor[HTML]{EAF7F0}+0.3} & 55.0 & \textsubscript{0.7} & {\cellcolor[HTML]{FEFDFD}--0.2} & 54.8 & \textsubscript{0.6} & {\cellcolor[HTML]{F1FAF5}+0.2} \\
\midrule
\midrule
& \multicolumn{3}{c}{\cellcolor[HTML]{3D85C6}{\color[HTML]{FFFFFF} \textit{100.0\%}}} & \multicolumn{3}{@{}c@{}}{\textit{0.0\%}}& \multicolumn{3}{@{}c@{}}{\textit{0.0\%}}& \multicolumn{3}{@{}c@{}}{\textit{0.0\%}} &&&& \multicolumn{3}{c}{\cellcolor[HTML]{3D85C6}{\color[HTML]{FFFFFF} \textit{100.0\%}}} & \multicolumn{3}{@{}c@{}}{\textit{0.0\%}}& \multicolumn{3}{@{}c@{}}{\textit{0.0\%}}& \multicolumn{3}{@{}c@{}}{\textit{0.0\%}}\\
\midrule
NER \multitask & 97.8 & \textsubscript{0.0} & {\cellcolor[HTML]{E0F3EA}+0.4} & 95.4 & \textsubscript{0.4} & {\cellcolor[HTML]{D8F0E4}+0.5} & 97.3 & \textsubscript{0.1} & {\cellcolor[HTML]{DEF2E8}+0.4} & 97.0 & \textsubscript{0.0} & {\cellcolor[HTML]{C7E9D8}+0.7} & 96.9 & \textsubscript{0.1} & {\cellcolor[HTML]{D3EEE1}+0.6} & 58.5 & \textsubscript{0.2} & {\cellcolor[HTML]{FEFEFE}--0.1} & 50.5 & \textsubscript{0.7} & {\cellcolor[HTML]{FEFBFB}--0.5} & 54.0 & \textsubscript{0.6} & {\cellcolor[HTML]{FEFAFA}--0.7} & 54.1 & \textsubscript{0.3} & {\cellcolor[HTML]{FDF8F7}--1.1} & 53.8 & \textsubscript{0.3} & {\cellcolor[HTML]{FEFAF9}--0.8} \\
NER \sequential & 97.3 & \textsubscript{0.6} & {\cellcolor[HTML]{FEFEFE}--0.1} & 95.1 & \textsubscript{0.6} & {\cellcolor[HTML]{F2FAF6}+0.2} & 96.6 & \textsubscript{0.5} & {\cellcolor[HTML]{FEFDFD}--0.3} & 96.6 & \textsubscript{0.5} & {\cellcolor[HTML]{EBF7F1}+0.3} & 96.4 & \textsubscript{0.5} & {\cellcolor[HTML]{FBFEFC}+0.1} & 57.7 & \textsubscript{1.2} & {\cellcolor[HTML]{FDF8F8}--0.9} & 50.5 & \textsubscript{2.0} & {\cellcolor[HTML]{FEFCFB}--0.5} & 52.6 & \textsubscript{1.1} & {\cellcolor[HTML]{FCF2F1}--2.0} & 54.4 & \textsubscript{0.9} & {\cellcolor[HTML]{FEFAF9}--0.8} & 53.5 & \textsubscript{1.0} & {\cellcolor[HTML]{FDF8F7}--1.1} \\
\bottomrule
\end{tabular}
}
\caption{
\textbf{Dialect-wise test results of the models trained on auxiliary tasks.}
The numbers in italics with blue backgrounds describe the dialect distributions in the data used to train the respective auxiliary tasks (e.g., 28.1\% of the training data for the syntactic tasks is in North Norwegian).
Key: 
\textit{B}\,=\,Bokmål,
\textit{N}\,=\,North N.,
\textit{T}\,=\,Trønder N.,
\textit{W}\,=\,West Norwegian,
$\Delta$\,=\,difference to the baseline model (in~pp.),
\multitask\,=\,multi-task learning,
\sequential\,=\,intermediate-task training.
}
\label{tab:auxtasks-dialects}
\end{table*}

\paragraph{All}
Table~\ref{tab:results-all} shows the development and test scores of all models (described throughout~\S\ref{sec:results}).

\paragraph{Baselines}
Table~\ref{tab:results-baselines-slots-detailed} provides the results of additional slot-filling metrics for the baselines~(\S\ref{sec:results-baselines}).

\paragraph{Noise}
Table~\ref{tab:noise-correlations} shows the correlations between the split word ratio difference of the noised training sets and the dialectal evaluation sets (cf.~\S\ref{sec:results-noise}).

\paragraph{Auxiliary tasks}
The remaining tables provide additional details for~\S\ref{sec:results-auxtasks}.
Table~\ref{tab:auxtask-details} focuses on the set-ups with auxiliary tasks and shows the scores on these tasks in addition to the SID scores.
Table~\ref{tab:auxtasks-dialects} focuses on the models trained on auxiliary tasks and shows the dialect distributions in the auxiliary task training data as well as the dialect-wise SID results.

\end{document}